\documentclass[conference]{IEEEtran}
\usepackage{times}

\usepackage[numbers]{natbib}
\usepackage{multicol}
\usepackage[bookmarks=true]{hyperref}
\usepackage{notoccite}
\usepackage{graphicx}
\usepackage{braket}
\usepackage{textcomp}
\usepackage{xcolor}
\usepackage{subcaption}
\usepackage{float}
\usepackage{multicol}
\usepackage{caption}
\usepackage{soul}
\usepackage{multirow}
\usepackage{array}
\usepackage{makecell}
\usepackage{hyperref}
\usepackage{url}
\usepackage{amssymb}
\usepackage{amsmath}
\usepackage{amsfonts}
\usepackage{booktabs}
\usepackage{algorithmic}

\usepackage[T1]{fontenc}
\usepackage[utf8]{inputenc}
\usepackage{babel}
\usepackage[font=small,labelfont=bf]{caption}

\newcommand{\deltahandnospace}{\textsc{DeltaHands}}
\newcommand{\telehandnospace}{TeleHand}

\newcommand{\telehand}{TeleHand\space}

\newcommand{\deltahands}{\textsc{DeltaHands}\space}
\newcommand{\Wtext}[1]{\textcolor{black}{#1}}

\begin{document}

\makeatletter
\let\@oldmaketitle\@maketitle
\renewcommand{\@maketitle}{\@oldmaketitle
  \includegraphics[width=\linewidth]
    {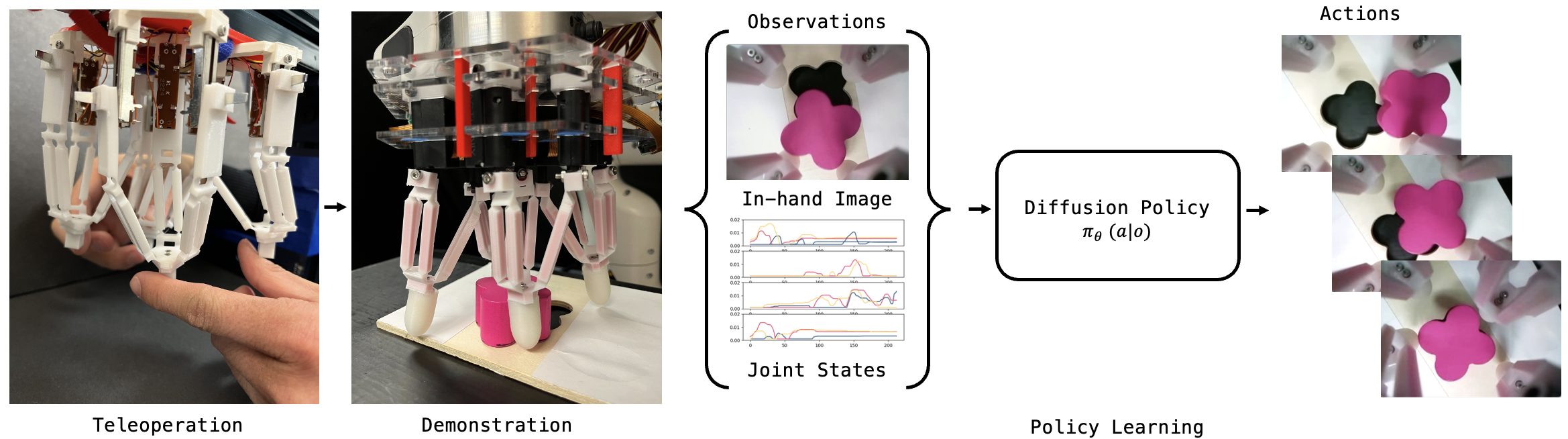}
  \captionof{figure}{$\widetilde{\mathit{Tilde}}$: \textit{$\underline{T}$eleoperation for Dexterous $\underline{I}$n-Hand Manipulation $\underline{L}$earning with a $\underline{De}$ltaHand}. We introduce an imitation learning-based in-hand manipulation system with a dexterous DeltaHand. We present a kinematic twin teleoperation interface,~\telehandnospace, to collect demonstrations on seven dexterous manipulation tasks, such as shape insertion shown above. By using vision-conditioned diffusion policies, the DeltaHand can autonomously complete the tasks.}
  \label{fig:teaser}
}
\makeatother

\title{$\widetilde{\mathit{Tilde}}$: Teleoperation for Dexterous In-Hand Manipulation Learning with a DeltaHand}


\author{\IEEEauthorblockN{Zilin Si$^{*}$, Kevin Zhang$^{*}$, Zeynep Temel, and Oliver Kroemer}
\authorblockA{$^{*}$Equal contribution \\
Robotics Institute, Carnegie Mellon University\\
\tt \{zsi,klz1,ztemel,okroemer\}@andrew.cmu.edu}
\url{https://sites.google.com/view/tilde-}
}

\maketitle
\begin{abstract}

Dexterous robotic manipulation remains a challenging domain due to its strict demands for precision and robustness on both hardware and software. While dexterous robotic hands have demonstrated remarkable capabilities in complex tasks, efficiently learning adaptive control policies for hands still presents a significant hurdle given the high dimensionalities of hands and tasks. To bridge this gap, we propose \textit{Tilde}, an imitation learning-based in-hand manipulation system on a dexterous DeltaHand. It leverages 1) a low-cost, configurable, simple-to-control, soft dexterous robotic hand, DeltaHand, 2) a user-friendly, precise, real-time teleoperation interface, \telehandnospace, and 3) an efficient and generalizable imitation learning approach with diffusion policies. Our proposed~\telehand has a \Wtext{kinematic twin} design to the DeltaHand that enables precise one-to-one joint control of the DeltaHand during teleoperation. This facilitates efficient high-quality data collection of human demonstrations in the real world. To evaluate the effectiveness of our system, we demonstrate the fully autonomous closed-loop deployment of diffusion policies learned from demonstrations across seven dexterous manipulation tasks with an average 90\% success rate.
\end{abstract}

\IEEEpeerreviewmaketitle

\section{Introduction}
\label{sec:intro}

Dexterous manipulation is essential for a wide range of real-world tasks such as inserting small components precisely for manufacturing, administering medicine in hospitals, and handling delicate ingredients while cooking.
However, a significant skill gap exists between human and robotic proficiency due to the demands for precision, robustness, and rapid adaptation to unstructured environments on both the hardware and software.
Take the in-hand shape insertion task (Fig.~\ref{fig:teaser}) as an example. The robotic hand has to adjust its control policy based on real-time sensory feedback, such as visual observations of the object, and seamlessly switch between skills like translation, rotation, and finger gaiting to align and insert the block into the template. Completing such high-dimensional and long-horizon tasks requires precise and dexterous manipulators as well as adaptable and robust policies that can handle diverse scenarios. Thus, integrated systems are necessary to address the challenges of dexterous manipulation and advance the field. 

Recent advances in imitation learning have shown great advantages in utilizing diffusion models~\cite{chi2023diffusion, reuss2023goal, wang2022diffusion} for efficient manipulation policy learning, as compared to deep reinforcement learning~\cite{ahn2020robel, andrychowicz2020learning} which is computationally expensive and data-hungry, or motion planning~\cite{cheng2023enhancing, liang2023learning, morgan2022complex} which relies on accurate modeling. 
However, imitation learning methods require high-quality demonstrations, which are challenging to collect quickly and reliably for dexterous manipulations. To leverage imitation learning, we need highly precise and easy-to-use teleoperation interfaces for dexterous robotic hands that will allow us to collect diverse demonstrations. 

Although anthropomorphic hands~\cite{shadowhand, puhlmann2022rbo, allegrohand, shaw2023leap} have already shown their ability to perform various manipulation tasks through teleoperation, these hands are designed to be general-purpose replacements for human hands which may be unnecessarily complex for certain domains. By contrast, non-anthropomorphic hands~\cite{ma2017yale, mccann2021stewart, si2024deltahands}, with their lower control complexity and higher design flexibility, can be better tailored to tasks such as precise peg insertion or in-hand manipulation. However, these designs present additional challenges for imitation learning given the human-to-robot hand correspondence problem.~\deltahands~\cite{si2024deltahands}, as shown in Fig.~\ref{fig:teaser}, are soft, compact, easy to customize, and possess high degrees-of-freedom (3-DoF per finger) that are simple to control, which makes them a great fit for dexterous in-hand manipulation. 
However, we need an intuitive and precise teleoperation interface to enable efficient imitation learning for~\deltahandnospace.

In this work, we present \textit{Tilde}, an imitation learning-based dexterous in-hand manipulation system built upon \deltahands~\cite{si2024deltahands} (Fig.~\ref{fig:teaser}). We first introduce a customized DeltaHand with an integrated in-hand camera for visual feedback and a novel 3D-printed finger design with hybrid soft and rigid materials that is capable of exerting on average 40\% more force than the original design with pure soft material. \Wtext{We then present a kinematic twin teleoperation interface named \telehandnospace, which has the same kinematics as the DeltaHand to enable a one-to-one mapping between joint states from the \telehand to the DeltaHand. This direct mapping allows for real-time, precise control of the DeltaHand. 
The \telehandnospace's user-friendly design simplifies human operation, making it an efficient tool for collecting high-quality human demonstration data.} Finally, by leveraging diffusion policies~\cite{chi2023diffusion}, we demonstrate the fully autonomous closed-loop deployment of our system on seven dexterous manipulation tasks including grasping, in-hand object rotations and translations using finger gaiting, precise shape insertion, and syringe pushing, all with an average success rate of 90\%. We believe that our low-cost and user-friendly integrated system can serve as a useful research platform for learning data-efficient dexterous in-hand manipulation policies. 

\section{Related Work}
\label{sec:related-work}
\subsection{Robotic Hands for Robot Learning}
Robot learning with robotic hands has been broadly studied for robotic manipulation~\cite{kroemer2021review}. Anthropomorphic hands such as the commercialized Shadow Hand~\cite{shadowhand} and Allegro Hand~\cite{allegrohand} have been wildly used for manipulation learning~\cite{arunachalam2023dexterous, handa2020dexpilot, qin2022one}, but they are intended for more general purpose usage and being close-sourced makes it difficult to adapt them to specific requirements. Open-sourced research hands such as the Leap Hand~\cite{shaw2023leap} provide research opportunities on robot learning with low-cost hardware. Alternatively, non-anthropomorphic hands have advantages in design and control simplicity. The Yale OpenHand project~\cite{ma2017yale} includes a series of hand designs that allow for fast prototyping and easy modifications, but they were optimized for simpler control with lower dexterity. ROBEL~\cite{ahn2020robel} was introduced as a platform for reinforcement learning benchmarks. \deltahands\cite{si2024deltahands} strike a good balance between cost, high dexterity, flexible design, and accurate yet simple kinematics which we adapt and customize in our integrated system. 

\subsection{Teleoperation Systems for Dexterous Manipulation}
Teleoperation for robots enables efficient data collection of human demonstrations for robot learning. \Wtext{Kinesthetic teaching~\cite{akgun2012keyframe, mulling2013learning} has been widely used to directly manipulate robots that have backdrivable motors and record the robot's joint positions. However, the human may occlude portions of the view when they move the robot, which is unsuitable for training control policies using visual feedback. In addition, kinesthetic teaching on soft robots may introduce motions from undesired and unrepeatable bending or buckling due to their compliant structure, instead of reflecting changes in the actuator's joint positions.}
Visual tracking has predominantly been used to collect low-cost demonstrations directly from human hand motions~\cite{arunachalam2023dexterous, handa2020dexpilot, qin2022one}. However, unstable, noisy visual tracking may require additional data smoothing processing or lead to more difficult policy training. This approach is also largely limited to anthropomorphic hands. Touch screen~\cite{toh2012dexterous}, and virtual reality (VR) setups~\cite{mannam2023framework,zhang2018deep} were proposed in recent years, but they suffer from human-to-robot hand correspondence issues. Custom teleoperation interfaces such as tactile gloves~\cite{amor2012generalization} for anthropomorphic hands and twin robot hardware ~\cite{wu2023gello, zhao2023learning} for robot arms with parallel-jaw grippers were presented as precise, intuitive, yet low-cost hardware. Similarly, we develop \telehandnospace, a kinematic twin teleoperation interface for a dexterous DeltaHand that is easy and intuitive to use.

\subsection{Learning for Dexterous Manipulation}
Reinforcement-learning in both simulation~\cite{andrychowicz2020learning, chen2023visual_dexterity,levine2016end, qi2023hand} and the real-world~\cite{ahn2020robel, kannan2023deft} have shown promising results in dexterous manipulation tasks such as solving in-hand Rubik's cubes and in-hand reorientation of novel objects. However, the huge exploration space results in heavy computation and data-generation costs which may be unsuitable for scaling up dexterous manipulation. Imitation learning, alternatively, has shown better sample efficiency while still possessing high performance~\cite{florence2022implicit, gupta2016learning, jain2019learning}, but it requires high-quality expert demonstrations. 
We utilize imitation learning to train policies from real-world human demonstrations. Specifically, we employ Diffusion Policy~\cite{chi2023diffusion}, which leverages diffusion models to handle the high dimensional task spaces and multi-modal action sequences of dexterous manipulation tasks. While most imitation learning suffers from domain shifts in data distribution, DAgger~\cite{ross2011reduction} was proposed to overcome this issue by using additional on-policy interactions and expert corrections. We incorporate this approach into our system to improve performance.
\section{Methodology}
\label{sec:method}

\begin{figure*}[ht]    
    \centering
    \includegraphics[width=0.95\linewidth]{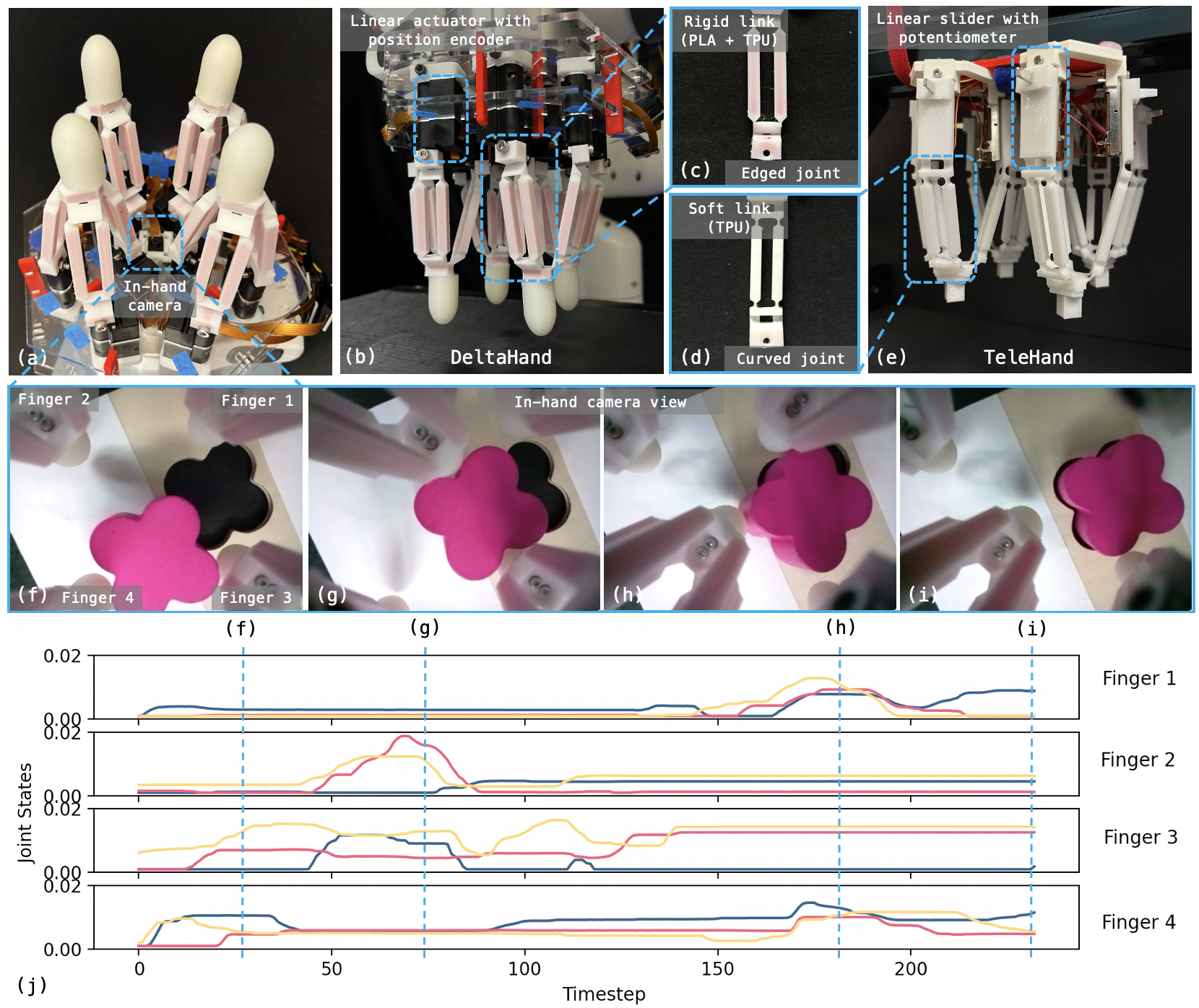}
    \caption{(a) A DeltaHand with an in-hand RGB camera. A kinematic twin teleoperation interface including (b) a DeltaHand and (e) a TeleHand. The TeleHand uses linear sliders with potentiometers to record the joint states of each finger. The DeltaHand will reproduce the motions of a TeleHand by using the Telehand's potentiometer readings as desired joint positions for its linear actuators. (c) The DeltaHand's fingers have 3D-printed rigid-core embedded links and edged joints, which increase the stiffness of each finger and enable them to exert more force. (d) The TeleHand's fingers have 3D-printed soft links and curved joints, which induce more compliance in each finger. Therefore less force is required for users to teleoperate the robot, which makes teleoperation easier. (f)-(i) In-hand camera images that capture the object and the DeltaHand's fingers. \Wtext{(j) The TeleHand's joint states indicate the movement of each finger during a demonstration.}}
    \label{fig:hand}
    \vspace{-2mm}
\end{figure*}
We present a dexterous manipulation learning system from three aspects: a dexterous robotic hand adapted from~\deltahandnospace, a kinematic twin teleoperation interface, and imitation learning with diffusion policies. We demonstrate key features of our system including 1) the high dexterity and precision of the DeltaHand, 2) the low latency, ease-of-use, and precision of the teleoperation interface,~\telehandnospace, and 3) the efficiency and generalizability of the policy learning for dexterous manipulation.

\subsection{Dexterous Hand Design}

\paragraph{Finger Design} \deltahands~\cite{si2024deltahands} is a configurable, highly dexterous, low-cost robotic hand framework based on Delta robots. The original DeltaHand's fingers were configured with 3D-printed soft TPU links (Fig.~\ref{fig:hand}(d)) for compliant and safe interactions. However, to benefit from the DeltaHand's compliance and extend its capabilities to manipulation tasks that require more forces such as pushing the plunger of a syringe, we modify the design of the Delta finger's links and joints as shown in Fig.~\ref{fig:hand}(c). 
To improve the force profile of the Delta finger, we 3D-print hybrid TPU and PLA links where we embed rigid PLA material (red) inside a thin outer shell of TPU (white). This strengthens the whole finger structure and enables fingers to apply more force while still preserving enough compliance to safely handle collisions. 
To improve kinematic precision, we use edged joints with smaller joint lengths instead of the original curved joints to reduce undesired buckling and deformations during finger motions. 
We conduct similar kinematics and force profile characterizations as~\cite{si2024deltahands} and observe that with the new design, the Delta finger's averaged kinematics error is reduced from $0.65$ mm to $0.53$ mm in position and $3.33$ degrees to $2.50$ degrees in orientation. In addition, the averaged lateral force profile is increased from $3.16$ N to $4.51$ N with $10$ mm actuation. 
These modifications can be easily incorporated by just swapping out the Delta fingers from the linear actuators, showcasing the flexibility of the~\deltahands framework.

\paragraph{In-hand Camera}Sensors are key components for closed-loop control by providing real-time feedback. In particular, local visual sensing is crucial for dexterous manipulation by capturing detailed geometric features~\cite{spector2021insertionnet}. Therefore, we integrate a mini Arducam camera module\footnote{\href{https://www.arducam.com/product/arducam-raspberry-pi-5mp-spy-camera-b0066/}{https://www.arducam.com/product/arducam-raspberry-pi-5mp-spy-camera-b0066/}} into the hand for in-hand visual sensing as shown in Fig.~\ref{fig:hand}(a). The camera can stream $640 \times 480$ resolution RGB images at $30$ fps over Wi-Fi using a Raspberry Pi 4. The camera is located at the center of the hand and on the same level as the Delta finger bases without taking extra space. It has symmetric observations to provide useful inductive bias~\cite{hsu2022vision}. We manually tune and fix the camera's focal length to focus on the area around the fingertips. The DeltaHand's kinematics permit a mostly unobstructed view of the fingertips which benefits visual servoing.

\paragraph{Fingertip Design} To increase the contact friction and enable soft contact for more secure grasps, we first 3D print the "bone" of the fingertip with TPU material, and then cast an additional layer of silicon rubber using Ecoflex 00-20 FAST\footnote{\href{https://www.smooth-on.com/products/ecoflex-00-20-fast/}{https://www.smooth-on.com/products/ecoflex-00-20-fast/}}.

\paragraph{Hand Configurations} An overview picture of the DeltaHand can be seen in Fig.~\ref{fig:hand}(a). We arrange four 3-DoF Delta fingers in a circular layout with a $40$ mm radius from the hand center to each Delta finger center. Each finger has a $40$ mm link length and $20$ mm base radius, and is individually actuated by three linear motors with $20$ mm stroke length. This gives a total of $12$ DoF and a $110$ mm × $110$ mm × $30$ mm workspace for the DeltaHand.

\subsection{Teleoperation Interface}

Previous work on \deltahands \cite{si2024deltahands} utilizes a Leap Motion camera for teleoperation.
When we conducted preliminary experiments with it, we found the visual tracking to be noisy and unstable, especially when fingertips are close together due to potential occlusions. 
In addition, since the Leap Motion was placed on the table surface and teleoperators held their hands in the air, the operators' hands would unavoidably drift over time because of fatigue which would inject additional noise into demonstrations. Given these reasons, we develop a kinematic twin teleoperation system for the DeltaHand to get precise and high-quality demonstrations.
The system includes a \telehand (Fig.~\ref{fig:hand}(e)) manipulated by a human teleoperator, and a DeltaHand (Fig.~\ref{fig:hand}(b)) to reproduce the \telehandnospace's finger motions in real-time.

The \telehand has the same configurations including the hand size, finger arrangement, and finger size, as the DeltaHand to enable direct one-to-one joint position mapping. The DeltaHand's fingers are actuated by linear motors with $20$ mm stroke length and each finger's link bases move prismatically. Similarly, the \telehand consists of linear sliders with a $20$ mm motion range to create the same linear mobility for each finger as the DeltaHand except they move passively. Teleoperators can easily drag and move the finger end-effectors of the \telehand which will lead to joint position changes in the sliders. The joint positions of \telehandnospace's fingers will be recorded by each sliders' potentiometers and then directly mapped to the DeltaHand as the linear motors' desired positions. For the \telehandnospace, we use the original Delta finger design (Fig.~\ref{fig:hand}(d)) which is more compliant and easier for humans to manipulate.

To enable real-time interfacing, we use Robot Operating System (ROS) for communication. Both the \telehand and the DeltaHand use Arduino microcontrollers to directly publish and receive ROS topics via a control PC. This also allows our teleoperation system to be easily integrated with other robotic arms. 
Our teleoperation system including the DeltaHand and the TeleHand can be manufactured in a day with off-the-shelf materials, 3D printing, and laser cutting, and costs around $\$1000$. From Fig.~\ref{fig:hand}(j), we show that the joint states read from the TeleHand are continuous and smooth which improves the training stability and efficiency.

\subsection{Learning with Diffusion Policies}
Diffusion models have shown their advantages while being used for policy learning from demonstrations~\cite{chi2023diffusion, reuss2023goal, wang2022diffusion} compared to behavior cloning~\cite{florence2022implicit, pomerleau1988alvinn}. They can greatly improve performance by capturing multimodal action distributions, and high-dimensional action spaces, which are key challenges for dexterous manipulation tasks. Therefore, we adapt the CNN-based Diffusion Policy~\cite{chi2023diffusion} to our system for dexterous in-hand manipulation policy learning with a DeltaHand. We condition the diffusion policies on visual observations from the in-hand camera and joint states of the DeltaHand (Fig.~\ref{fig:hand}(f)-(j)) and predict action sequences. Both the joint states and actions are represented as the 12-dimensional absolute actuator joint positions. \Wtext{We trained policies using either actuator joint positions or end-effector positions derived from forward kinematics as inputs, and we experimentally found that using joint positions resulted in better performance.}

Imitation learning methods for long-horizon tasks are known to suffer from covariate shift ~\cite{ross2011reduction}.
To avoid this issue, we use DAgger~\cite{ross2011reduction} to add on-policy interactions if the policy fails during the inference, and refine the policy with these DAgger demonstrations.
In addition, data augmentation has been broadly used to improve the generalization and robustness of learning, especially for dexterous manipulation tasks with high dimensional state and action spaces. \Wtext{Through experimentation, we found that using various data augmentation techniques on observations greatly improved task performance. We leveraged 1) random image cropping and rotation to improve the rotational and translational invariance of fingers' visual servoing to the objects, and 2) Gaussian noise to joint state observations to guide the policy in learning funneling behaviors that can make the policy more robust when encountering unseen joint states. Specifically, we randomly cropped the images from their original size of $(240, 320)$ to $(216, 288)$ and rotated the images within 30 degrees. We added Gaussian noise with a standard deviation of 3.16 mm to each joint state. However, we found that random resized cropping does not improve performance because the camera is fixed w.r.t. the hand. In addition, some tasks are directional and the fingers' movements are reflected in the in-hand view, so directional invariance created by image flipping is therefore detrimental. }
\section{Experimental Setup}
\label{sec:exp}

We evaluate our proposed system on seven dexterous manipulation tasks as shown in Fig.~\ref{fig:tasks}: \textbf{Grasp}, \textbf{Block Slide}, \textbf{Block Lift}, \textbf{Ball Roll}, \textbf{Cap Twist}, \textbf{Syringe Push}, and \textbf{Shape Insert}. \textbf{Grasp} is a fundamental skill for most manipulation tasks. The second to fifth tasks focus on different in-hand object repositioning skills: \textbf{Block Slide} corresponds to horizontal XY translations, \textbf{Block Lift} corresponds to vertical Z translations, \textbf{Ball Roll} corresponds to rotations around the X and Y axes, and \textbf{Cap Twist} corresponds to rotations around the Z axis. The above tasks mostly require repeated motions, while the last two tasks consist of multi-modal action sequences and longer time-horizons. \textbf{Syringe Push} requires fingers to precisely align the syringe before pushing the plunger, while \textbf{Shape Insert} requires the fingers to translate, rotate, and transport the object to the final goal pose. Through our experiments, we show the capability of our proposed system to handle these dexterous manipulation tasks. 

\subsection{Data Collection}

\begin{figure}    
    \centering
    \includegraphics[width=0.98\linewidth]{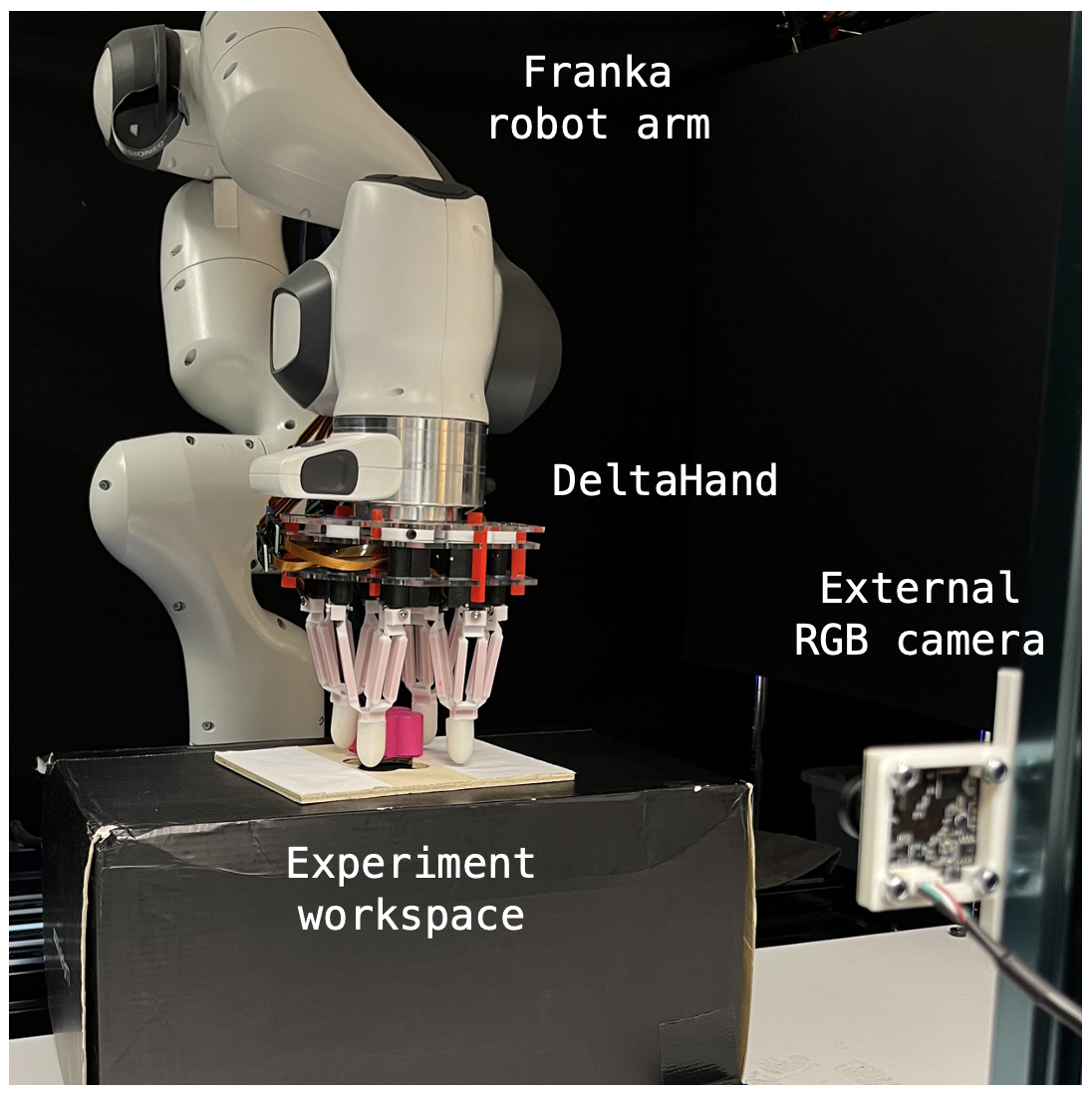}
    \caption{Experimental setup. We mount a DeltaHand on a Franka robot arm. We pre-set the height and location of the Franka arm on top of the experiment workspace. An external RGB camera is mounted in front of the experiment workspace.}
    \label{fig:exp-setup}
\end{figure}

We mount a DeltaHand on a Franka robot arm as shown in Fig.~\ref{fig:exp-setup}. For most tasks, we keep the robot arm static while the DeltaHand uses its in-hand capabilities to manipulate the objects. An external RGB camera\footnote{\href{https://www.amazon.com/gp/product/B0C289GYVZ/}{https://www.amazon.com/gp/product/B0C289GYVZ/}} is placed in front of the experiment workspace. For each task, we manually preset the height and the location of the arm to approximately align the DeltaHand's workspace with the object. To collect demonstrations, we first define the goal for each task which can be verified from the visual observations. If we reach the goal, we end the demonstration, or we run until we reach 5000 time steps which roughly equates to 250 seconds (data collection runs at 20fps speed). \Wtext{To collect DAgger demonstrations during inference, we first deploy the learned policy and let the DeltaHand manipulate an object autonomously while a human monitors. When the human decides that the policy has failed or is unlikely to finish the task, such as when the fingers stop or oscillate between similar states repeatedly, the human will pause the policy deployment. When the policy is paused, the human teleoperator first manually moves the fingers of TeleHand to match the current DeltaHand's finger positions to reproduce the failure configuration and then takes over the control of the DeltaHand with the Telehand to finish the task with teleoperation. We use these teleoperated DAgger demonstrations from the failure point to fine-tune the policy.}

\subsection{Training Details}
We train CNN-based Diffusion Policies~\cite{chi2023diffusion} for all tasks. For visual features, we use ResNet18 as the visual encoder for both in-hand and external observations and then concatenate the 512-dimensional visual features with the 12-dimensional joint state vector. Actions are represented as the next timestep's joint state. We fix the observation horizon, action prediction horizon, and action execution horizon to 2, 16, and 8 for all tasks, respectively. We train each policy model for 100 epochs with the AdamW optimizer using learning rate=$1e-4$ and weight decay=$1e-6$.

To improve the training stability, we normalize all the joint states and actions to the range $[-1, 1]$, and images to $[0, 1]$. In-hand images, external images, and joint states from demonstrations are synchronized to 20 fps. However, we found that down-sampling the data by a factor of 3 reduces the effect of the idle actions from demonstrations.

\begin{figure*}
    \centering
    \includegraphics[width=0.80\linewidth]{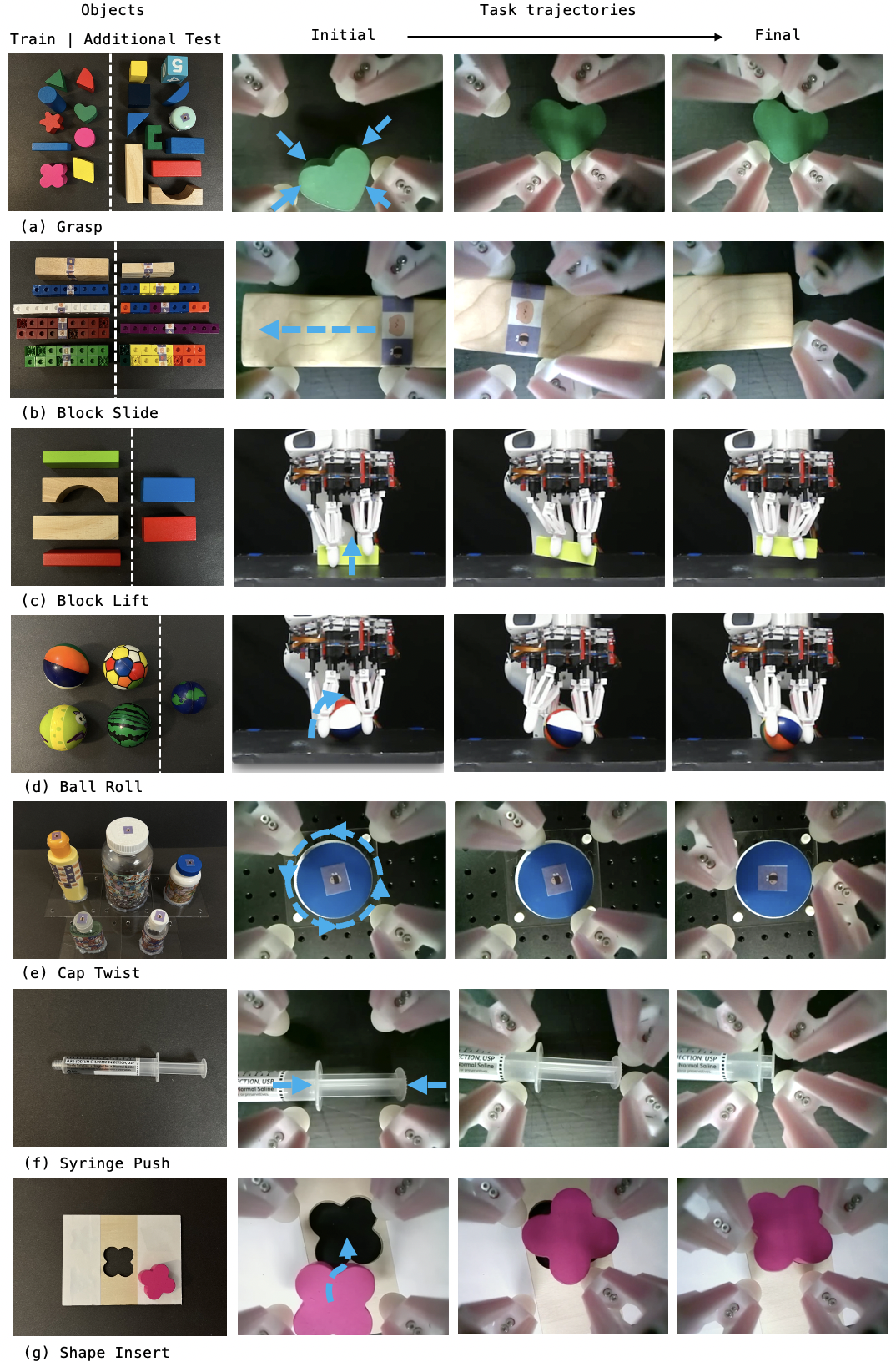}
    \caption{Task gallery. We evaluate our system on seven dexterous manipulation tasks: (a) \textbf{Grasp} (b) \textbf{Block Slide} (c) \textbf{Block Lift} (d) \textbf{Ball Roll} (e) \textbf{Cap Twist} (f) \textbf{Syringe Push}  (g) \textbf{Shape Insert}. The goals of tasks are indicated by blue arrows in the initial images of task trajectories. For tasks (a)- (d), we separate the training and additional unseen testing objects with white dashed lines.}
    \label{fig:tasks}
\end{figure*}

\subsection{Tasks}
An overview of all tasks is shown in Fig.~\ref{fig:tasks}. We aim to use these tasks to evaluate the policies' capabilities, efficiency, and generalizability. For tasks (a) - (d), we evaluate the policies with additional unseen test objects as displayed on the right side of the object sets. For tasks (f) and (g), we randomly initialize the objects within the experiment workspace.

\begin{table*}[t]
\centering
\setlength{\tabcolsep}{4pt}
\renewcommand{\arraystretch}{1.2}
\begin{tabular}{lccccccc}
\Xhline{2\arrayrulewidth}
\textbf{Task} & \textbf{Grasp} & \textbf{Block Slide} & \textbf{Block Lift} & \textbf{Cap Twist} & \textbf{Ball Roll} & \textbf{Syringe Push} \\
\hline
{\# demos} & 45 & 40 & 20 & 30 & 20 & 50 \\ 
{\# DAgger demos} & 10  & 0 & 10 & 10 & 5  & 20 \\
\hline
{\# Success / \# tests before DAgger} & 17/20 &  10/10  & 7/10  & 8/10 &  7/10 & 6/10 \\ 
{\# Success / \# tests after DAgger} & 20/20 &  10/10   & 8/10 & 10/10 & 10/10  & 8/10 \\
\Xhline{2\arrayrulewidth}
\end{tabular}
\caption{Experimental results on six tasks. We show that with less than 50 demos, we can achieve success rates over 60\% on all tasks before DAgger. With additional DAgger demonstrations, all tasks have improved results and achieved success rates over 80\%.}
\label{table:results}
\end{table*}

\textbf{Grasp} We grasp a set of objects with various shapes, colors, and weights. We first teleoperate the DeltaHand to center the object and then grasp the object using all fingers. Afterwards, we lift the robot arm $10$ cm to check whether the grasp is stable. 

\textbf{Block Slide} We slide rectangular blocks of different sizes and colors horizontally from right to left. We tape the center of the blocks to provide distinctive features and allow for consistent human resets. During the demonstrations, we teleoperate the right set of fingers to move the object while we position the left set of fingers to form a funnel to stabilize the block. We end the demonstration once the right end of the block is located in the center of the in-hand view. 

\textbf{Block Lift} We lift blocks of differing sizes, shapes, and colors up from the table by alternating grasps on the block with the right and left set of fingers. 

\textbf{Ball Roll} We roll the ball between the fingers. We use two opposing fingers to loosely hold the ball while the other two fingers roll the ball around the axis defined by the virtual connecting line of the two holding fingers. 

\textbf{Cap Twist} We rotate the cap of bottles 360 degrees. The bottom parts of the bottles are fixed to the table and the cap is twisted by two fingers. A full rotation is indicated by the bee sticker on the cap. 

\textbf{Syringe Push} We push the plunger to close the syringe. We place the left set of fingers on the syringe barrel while we position the right set of fingers at the end of the plunger. The right and left sets of fingers have to align with the syringe and then push together. Any misalignment during the push will block and jam the pushing which requires repositioning fingers. During initialization, the syringe is randomly placed between the fingers on the table and the plunger is opened to a random length. We removed the rubber tip of the plunger which usually seals in liquid to reduce the friction and forces for this set of experiments. We further evaluated the forceful syringe push with the rubber tip on the plunger in Section.~\ref{sec:challenging-task} as a challenging task.

\textbf{Shape Insert} We move and match a flower-shaped block to its corresponding template hole on a board. Both the board and the object are initialized randomly and the DeltaHand needs to translate, rotate, and transport the object to match the template hole. 

\begin{figure*}[t]   
    \centering
    \includegraphics[width=0.98\linewidth]{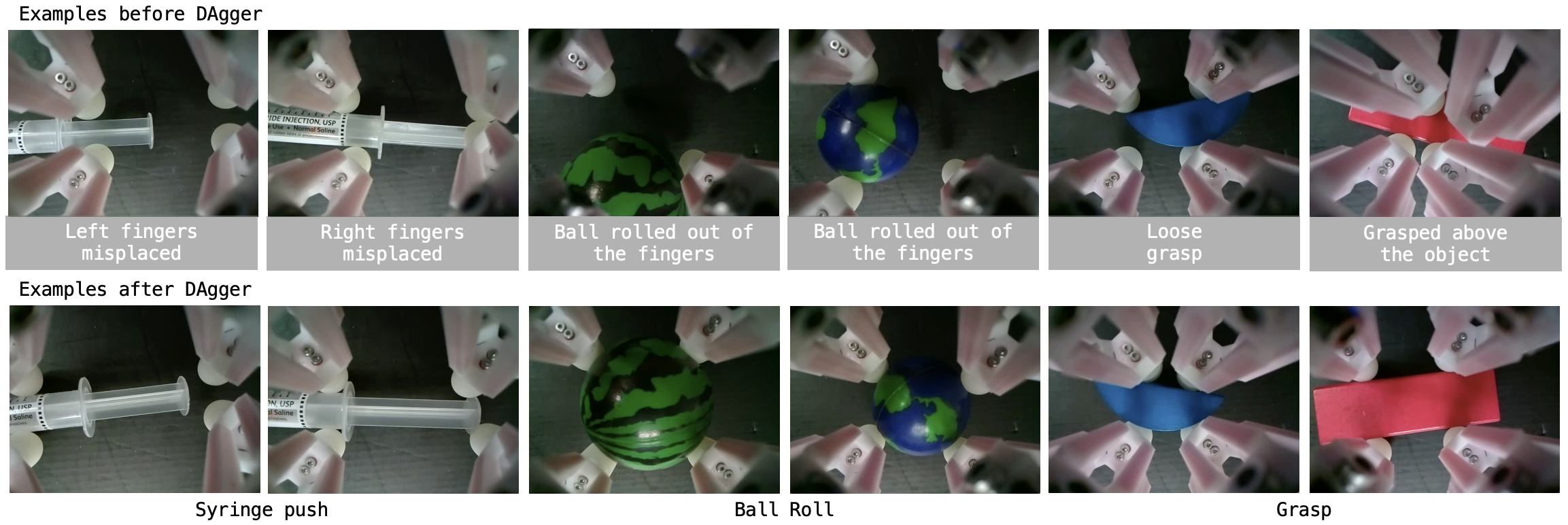}
    \caption{Qualitative comparisons between task executions from policies trained before and after DAgger demonstrations. By refining the policies with corrective demonstrations from failure cases, the policies can handle these challenging scenarios.}
    \label{fig:dagger-example}
\end{figure*}

\begin{figure*}[t] 
    \centering
    \includegraphics[width=0.98\linewidth]{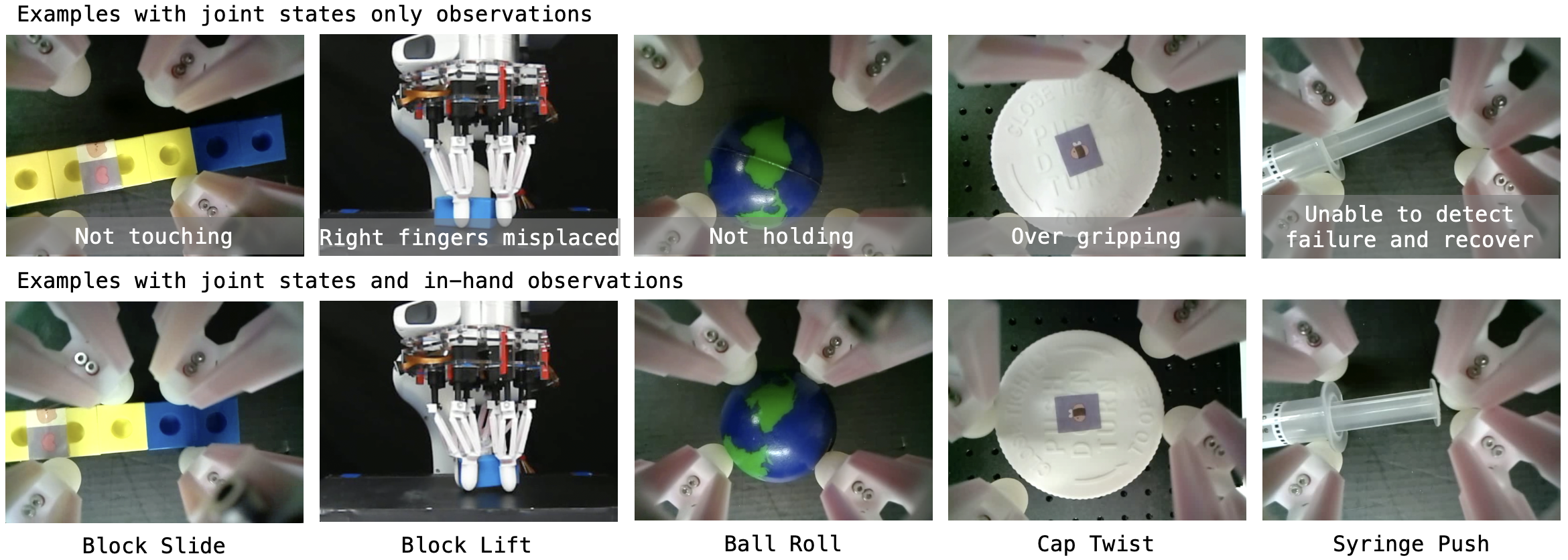}
    \caption{Qualitative comparisons between task policies conditioned on observations with joint states only versus joint states with in-hand images. We observe that visual observations improve the generalizability of the policies by adapting the contact points of the fingers to the objects' shapes, sizes, and locations.}
    \label{fig:observation-viz}
    \vspace{-3mm}
\end{figure*}

\begin{figure}[t]    
    \centering
    \includegraphics[width=0.98\linewidth]{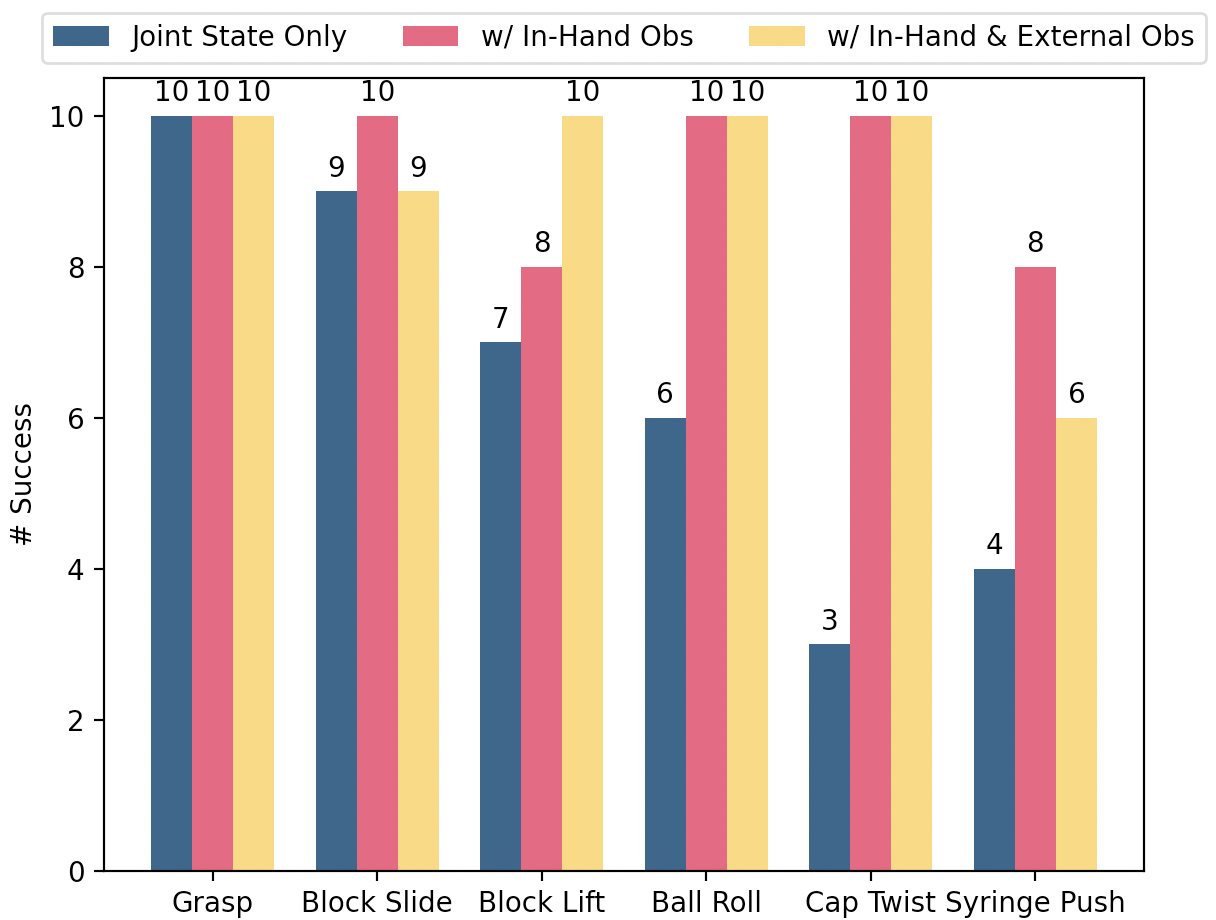}
    \caption{Effect of using visual observations. We evaluate the policies trained with observations as 1) joint states only, 2) joint states with in-hand images, and 3) joint states with in-hand and external images. The results show performance improvement with visual observations compared to only using joint states as observations.}
    \label{fig:ablation-sensors}
    \vspace{-3mm}
\end{figure}

\section{Experimental Evaluations}
We experimentally evaluate the performance of our proposed system on the aforementioned tasks, and we aim to answer the following questions: 
\begin{enumerate}
    \item How practical is our system at performing dexterous manipulation tasks?
    \item How effective are the in-hand observations on the task performance?
    \item How efficiently can our system learn robust policies with a limited number of demonstrations?
\end{enumerate}

\subsection{Experimental results}

The number of demonstration data for the first six tasks can be seen in the first and second rows of Table~\ref{table:results}. They vary depending on the task difficulty and the number of objects we use. After the initial policy training, we run 10 test trials on both the train and additional test objects. If the policy fails during the evaluation, we collect new demonstrations starting from the failure configuration and teleoperate the DeltaHand until success is achieved. These DAgger demonstrations are then used to refine the policy. The success rates before and after DAgger are shown in the third and fourth rows of Table.~\ref{table:results}. We use joint states and in-hand images as observations for all tasks. We observe that for most tasks, with less than 50 demonstrations, we can achieve a success rate of over 60\%, and even a 100\% success rate on the \textbf{Block Slide} task. For other tasks, DAgger demonstrations can improve the success rates to over 80\% and recover from previously encountered failure cases.

\subsection{Failure Analysis}
Fig.~\ref{fig:dagger-example} shows some common failure cases during the initial tests and how DAgger improves the performance on these.  In the initial \textbf{Syringe Push} evaluations, the left two fingers occasionally incorrectly grasped the plunger when the syringe was placed more to the left. This showed that the policy did not know to position the left two fingers onto the barrel of the syringe. With additional DAgger demonstrations, the policy was able to move the left two fingers further to the left when necessary to hold onto the barrel of the syringe while the right two fingers pushed the plunger. Another common failure case was that the right two fingers occasionally grasped onto the plunger instead of positioning themselves behind the end of the plunger because the syringe was initialized too far to the right. With DAgger, the policy was able to either make the two left fingers pull the syringe more to the left before the right two fingers started pushing on the plunger, or make the right two fingers first push the plunger as much as they could and then reposition themselves behind the end of the plunger and push the rest of the way in. These additional demos from the failure cases enabled the diffusion policies to learn more robust recovery behaviors.

In the \textbf{Ball Roll} task, the balls occasionally rolled out from the grasp of the two opposing fingers when pushed. By using additional DAgger examples, the DeltaHand was able to adapt to different sizes of balls and also stop rotating the ball when the policy determined the ball was not positioned correctly and reposition the ball before restarting to rotate it.

Finally, in the \textbf{Grasp} examples, the DeltaHand was unable to grasp some unseen objects that were of differing heights and shapes such as flat blocks versus tall blocks and semicircle blocks. With additional DAgger examples on these objects, the policy was able to successfully adapt to the new objects. We conclude that with more diverse data, the domain shifts from the training data distributions were reduced and the policies became more robust to different scenarios.

\begin{figure}[t] 
    \centering
    \includegraphics[width=0.90\linewidth]{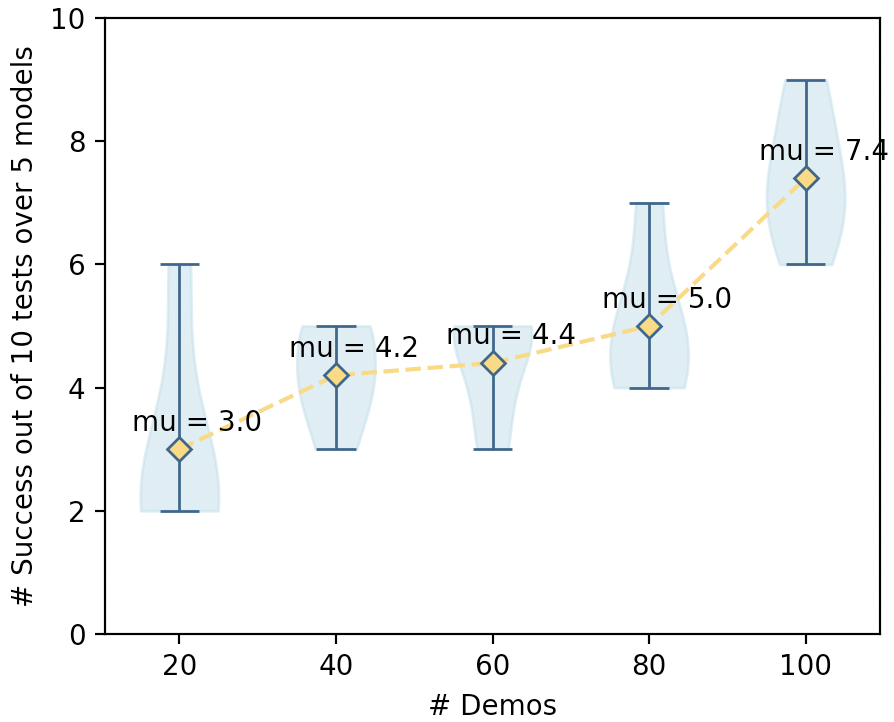}
    \caption{Data efficiency evaluation. We evaluate policies trained with 20, 40, 60, 80, and 100 demonstrations of the \textbf{Shape Insert} task and average the performances over five models using different random seeds. The success rates increased with more training data.}
    \label{fig:ablation-dataset}
    \vspace{-1em}
\end{figure}

\subsection{Effect of using visual observations}

We have two types of observations: DeltaHand's joint states from motor encoders as proprioception, and in-hand images or external images as exteroception. We evaluate how sensitive the tasks are to different kinds of observations. As shown in Fig.~\ref{fig:ablation-sensors}, we train the policies with observations as 1) joint states only, 2) joint states and in-hand images, and 3) joint states, in-hand images, and external images. We show that most tasks achieve better performance with in-hand images compared to using joint states only. However, with additional external images, the performance does not differ much from solely using in-hand images with joint states. We noticed that in most failure cases, the objects are hard to see from the external camera due to their small size and occlusions from the fingers, thus the external images are not as informative as the in-hand images. But for the \textbf{Block Lift} task, the policy with external images gets better results because the external camera can capture the object's state better such as the height of lifting, which is more informative. For the \textbf{Grasp} task, all three policies achieved a 100\% success rate and we give significant credit to the compliance of the DeltaHand which can passively adapt to most objects even without in-hand visual observations.

Some examples for comparison can be seen in Fig.~\ref{fig:observation-viz}. By using joint states only, common failure cases were that fingers would not contact the object and were unable to move or secure the object, they over-gripped the object which restricts movement, or were unable to detect failure and recover. By leveraging in-hand visual observations, the fingers can visual servo to adaptively make contact with objects based on their shapes, sizes, and locations.

\begin{figure}[t]
    \centering
    \includegraphics[width=0.98\linewidth]{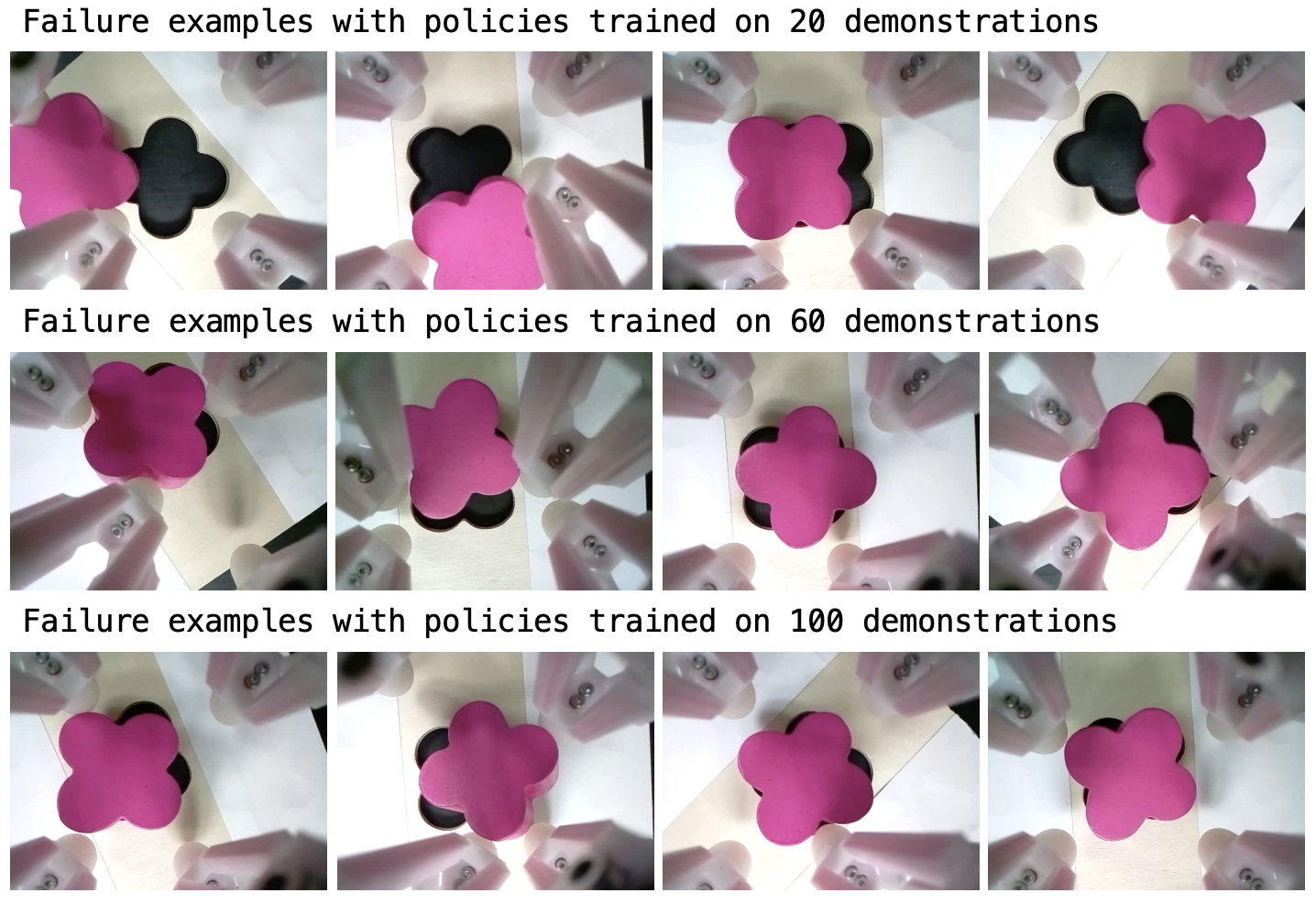}
    \caption{Examples for comparison between policies trained with 20, 60, and 100 demonstrations. \textbf{Shape Insert} requires the hand to transport and reposition objects to align with the template hole. With less training data, the task mostly failed at the early transportation stage. When the training data size increased, the policies often failed during the final fine orientation alignment stage where most cases are visually ambiguous.}
    \label{fig:flower-failure}
    \vspace{-0.5em}
\end{figure}

\begin{table}[t]
\centering
\setlength{\tabcolsep}{4pt}
\renewcommand{\arraystretch}{1.2}
\begin{tabular}{p{1.0cm}cccc}
\Xhline{2\arrayrulewidth}
\textbf{Task} & \textbf{Observations}& \textbf{BC~\cite{mandlekar2022matters}} & \textbf{IBC~\cite{florence2022implicit}} & \textbf{Diffusion Policy~\cite{chi2023diffusion}} \\
\hline
{\textbf{Cap}} & State & 0/10  & 0/10 & 3/10  \\
{\textbf{Twist}} & State + In-Hand & 4/10  & 0/10 & 10/10  \\
\hline
\multirow{2}{1.5cm}{\textbf{Shape Insert}} & State & 0/10 &  3/10   & 2/10 \\
 & State + In-Hand  & 2/10 &  3/10   & 7/10 \\
\Xhline{2\arrayrulewidth}
\end{tabular}
\caption{\Wtext{We evaluate the performance of three Imitation Learning methods: Behavior Cloning (BC) (Robotmimic~\cite{mandlekar2022matters}), Implicit Behavior Cloning (IBC)~\cite{florence2022implicit}, and Diffusion Policy~\cite{chi2023diffusion} on two tasks, \textbf{Cap Twist} and \textbf{Shape Insert}, either with joint states only or with joint states and in-hand visual observations. We show Diffusion Policy achieves the best performance.}}
\label{table:il-compare}
\vspace{-1.5em}
\end{table}

\subsection{Data Efficiency Evaluation}
From Table.~\ref{table:results}, we see that most tasks can achieve 100\% success rates with less than 50 demonstrations. These tasks such as \textbf{Block Slide}, \textbf{Block Lift}, \textbf{Cap Twist}, and \textbf{Ball Roll}, use repeated motion patterns. For long-horizon tasks, multi-modal actions are required to complete the tasks. For instance, in \textbf{Shape Insert}, the fingers need to rotate, translate, or transfer objects between different sets of fingers as shown in Fig.~\ref{fig:hand} (f)-(i). To evaluate data efficiency on the \textbf{Shape Insert} task, we randomly sample five sets of 20, 40, 60, and 80 demonstrations from the dataset of 100 demonstrations and train policies with these data subsets respectively. The evaluation results in Fig.~\ref{fig:ablation-dataset} show an increase in success rate when policies are trained on more data, indicating that the policy has better generalizability given more data. 

We visualize common failure cases from policies trained with 20, 60, and 100 demonstrations in Fig.~\ref{fig:flower-failure}. We observe that with less training data, the task failed mainly at the early transportation stage where the fingers are unable to move the object closer to the template hole. With more training data, the failures occurred closer to the final orientation alignment stage. We conclude that diffusion policies can learn more complex and finer manipulation skills with more data.

{\color{black}\subsection{Comparison of Different Imitation Learning Methods}
To support our design choice of using diffusion policies, we comparably evaluate three Imitation Learning methods: Behavior Cloning (BC) (Robotmimic~\cite{mandlekar2022matters}), Implicit Behavior Cloning (IBC)~\cite{florence2022implicit}, and Diffusion Policy~\cite{chi2023diffusion} either with joint states only or with joint states and in-hand visual observations on two tasks: \textbf{Cap Twist} and \textbf{Shape Insert}. We report the performance of each method in Table.~\ref{table:il-compare}. We show that the Diffusion Policy achieved the best performance. We observe that 1) BC can predict smooth trajectories due to the continuity of the MLP network and has a fast inference speed (60 FPS) with its small model scale. However, it fails to reason about the multi-modal actions necessary to complete more complex tasks such as \textbf{Shape Insert}. For the \textbf{Shape Insert} task, it often fails at the final fine orientation alignment stage. On the other hand, actions predicted from 2) IBC contain a lot of noise due to the sampling process, which is unsuitable for dexterous manipulation tasks that require high precision. The successful trials on the \textbf{Shape Insert} task resulted from random interactions with the block that did not reflect clear intentional trajectory motions. Finally, 3) Diffusion Policy can handle multi-modal actions while also being able to predict relatively smooth trajectories at a real-time inference speed (20 FPS). We include more implementation details of these three methods in Section~\ref{sec:policy-detail}.}

{\color{black}\subsection{Evaluation on Robustness and Generalization of Policies}

\begin{figure}[t]
    \centering
    \includegraphics[width=0.8\linewidth]{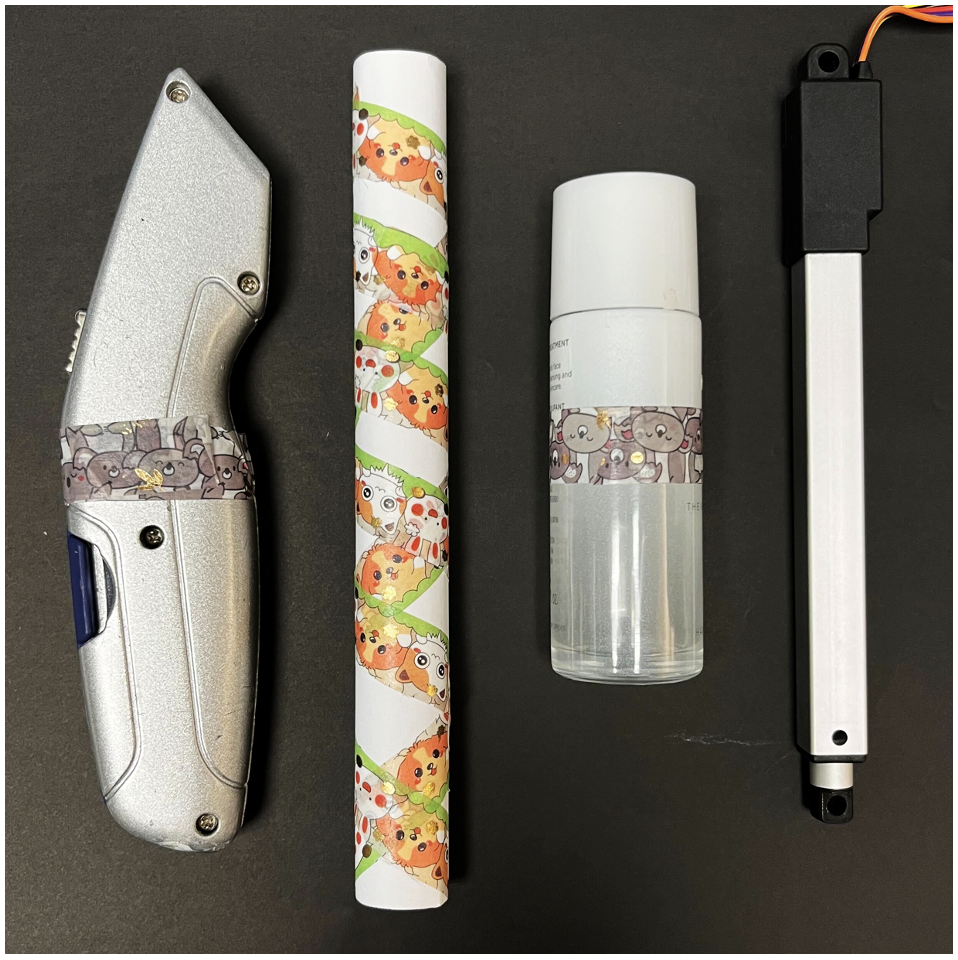}
    \caption{\Wtext{We evaluate the generalization of policies with additional unseen irregular-shaped objects used on the \textbf{Block Slide} task.}}
    \label{fig:generalize}
    \vspace{-5mm}
\end{figure}

The robustness and generalizability of policies to unstructured environments is crucial for real-world deployment. Along with the evaluation on unseen objects, we further evaluate the robustness of the learned policies to randomly initialized hand poses on the \textbf{Block Slide} task. For each trial, we first generate a random relative transformation consisting of a z-axis translation, and rotations around the x and y axes based on the original pre-defined fixed hand pose. The sampling range of the z-axis translation is between $(-5, 5)$ mm and $(-10, 10)$ degrees in both the x-axis and y-axis orientations. 
We then move the DeltaHand to this randomly initialized pose and begin policy inference. The policy is still able to succeed in $17$ of $20$ trials. In addition, we include several unseen irregularly-shaped objects in these 20 tests, as shown in Fig.~\ref{fig:generalize}, and the learned policy can still succeed without any additional training.
We observe that the compliance of the fingers can help with misalignments between the hand and the object and compensate for policy errors. Most failure cases occurred when the object shifted outside the hand's workspace due to the random initialization, and the fingers were no longer able to reach the object. This issue can be potentially addressed by integrating robot arm motions to adjust the hand pose in future work. 
}

\begin{figure*}[t] 
    \centering
    \includegraphics[width=0.98\linewidth]{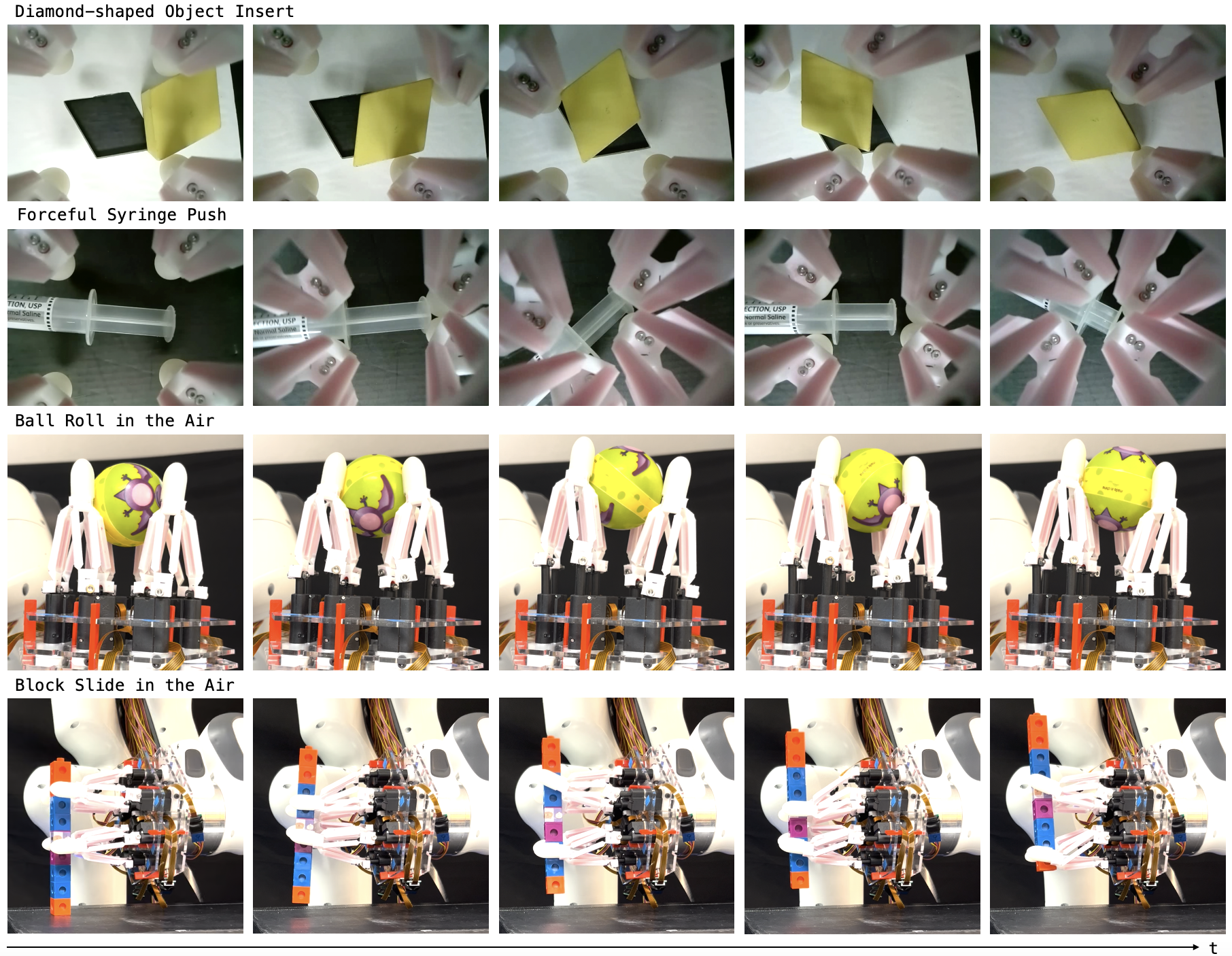}
    \caption{\Wtext{Test examples of four challenging tasks: diamond-shaped block insertion, forceful syringe push, ball roll in the air, and block slide in the air over time. Diamond-shaped block insertion requires orientation alignment of more than $90$ degrees. Forceful syringe push requires higher alignment precision to succeed. Both ball roll and block slide in the air have non-recoverable failure risks.}}
    \label{fig:challenging-task}
    \vspace{-3mm}
\end{figure*}

\subsection{Performance on Challenging Tasks}
\label{sec:challenging-task}
\paragraph{Diamond-shaped Block Insert} The flower-shaped block is radially symmetric and can be aligned with less than a 45-degree rotation. We further tested our system with a diamond-shaped block that requires up to 180 degrees of orientation alignment. The demonstration trajectory length of the diamond-shaped block insertion is on average 23.65 seconds compared to 18.08 seconds for the flower-shaped block insertion. We show that with 80 demonstrations, the learned policy can still handle this long-horizon task with 5 successful runs out of 10 trials. A test example can be seen in Fig~\ref{fig:challenging-task}. We show that the policy learns to coordinate movements among fingers to re-orient the object over 90 degrees. Most failures are due to unseen object poses, which we believe can be resolved with more data. 

\paragraph{Forceful Syringe Push} We further evaluate a forceful syringe push task by adding the rubber seal to the tip of the plunger where the friction between the plunger and the syringe barrel demands around 4.0 N to push the syringe. With 40 demonstrations, we can get 8 successful runs out of 10 tests. A test example can be seen in Fig~\ref{fig:challenging-task}. We observe that the policy can detect the misalignment caused by the pushing force in the third plot of Fig~\ref{fig:challenging-task} and realign the fingers with the syringe to complete the task.

{\color{black}\paragraph{Finger Gaiting in the Air}
In all the aforementioned tasks, objects are mainly supported on the table and can be recovered if the policies perform incorrect actions. We further evaluate two challenging finger-gaiting-in-the-air tasks including \textbf{Ball Roll} and \textbf{Block Slide} which have non-recoverable failure risks. Examples can be seen in Fig.~\ref{fig:challenging-task} where the fingers hold and rotate a ball upright for \textbf{Ball Roll} and pinch and slide a block upward for \textbf{Block Slide}. We quantitatively evaluated these two tasks and obtained 9 and 6 successful runs out of 10 tests respectively. The singular failure case for \textbf{Ball Roll} occurred when the ball fell down between fingers. This resulted in a large visual occlusion from the in-hand camera view which lead the policy to fail. Most failure cases for \textbf{Block Slide} resulted when one pair of fingers pinched too tightly and the other pair could not move the block any further. This demonstrates the importance of incorporating tactile sensors in the fingers which we plan to explore in future work.
}

{\color{black}\subsection{Limitations and Discussions}
From the experimental results, we demonstrate our system's capability on a varied set of grasping and in-hand translation and rotation manipulation tasks. Through our extensive evaluations, we observe the limitations of our system as follows:
\subsubsection{Lack of Tactile Sensing} We show that in-hand visual observations have greatly improved the performance of the learned policy from our ablation study. However, tactile sensing is crucial in providing direct force feedback for delicate object manipulation which have non-recoverable failure modes. Although we can still achieve many tasks by solely using visual feedback, the policy needs to learn more robust behaviors and relies on repeated attempts to achieve the target goal. Tactile feedback could lead to more intentional and controlled interactions with objects in highly-dexterous tasks.
\subsubsection{Generalization to Unstructured Environments} We show that the learned policies can adapt to unseen objects in a top-down tabletop environment. However, we observe failure cases with environments out of the training distribution such as different backgrounds or large disturbances in the hand poses. To improve the robustness of in-hand tasks, we plan to explore object-centric learning approaches such as conditioning the diffusion policy on object segmentation masks to better handle unseen environments and allow for explicit reasoning about objects in clutter near or in the hand. 
\subsubsection{Lack of Robot Arm Motion} We emphasize the in-hand manipulation capabilities of our proposed system in this work, and our evaluations focus on finger-based manipulations. However, integrating robot arm motions can improve robustness and extend the capabilities of our system beyond in-hand manipulation. For example, when randomly disturbing the hand pose, we observe failure cases of objects being outside of the fingers' workspace. Leveraging the robot arm's movement could address such limitations by properly initializing and aligning the hand's pose with the target object. This can also enable extrinsic dexterous manipulation.
}
\section{Conclusions} 
\label{sec:conclusions}
We present \textit{Tilde}, an imitation learning-based in-hand manipulation system with a dexterous DeltaHand. We introduce a kinematic twin teleoperation interface for low-cost data collection of high-quality human demonstrations and efficient end-to-end real-world policy learning by using diffusion policies. We show that with our system, we can perform a variety of dexterous manipulations and achieve an average success rate of 90\% across our evaluation tasks. These tasks include grasping, in-hand object re-positioning and re-orientation, and finger gaiting. Our experiments show the capability of the system to learn robust vision-based dexterous manipulation policies from demonstrations that were acquired with our easy-to-use and precise teleoperation interface.

\Wtext{In the future, we would like to improve the generalizability of our system for broader dexterous manipulation tasks in unstructured environments. Therefore, we plan to augment sensing modalities with tactile sensing by incorporating fingertip tactile sensors for more delicate tasks, integrate arm motions into our teleoperation system to achieve intrinsic and extrinsic dexterity~\cite{gupta2022extrinsic, zhou2023learning}, and explore object-centric approaches to improve the policies' robustness to unseen scenarios.}

\section*{Acknowledgments}
The authors would like to thank Sha Yi, Shashwat Singh, Jennifer Yang,  Moonyoung (Mark) Lee, Mohit Sharma, Saumya Saxena, and Haomin Shi for the their help in discussions, experiments, and with manuscript revisions. This work was supported by the National Science Foundation under grant No. CMMI-2024794 and Google. 


\bibliographystyle{plainnat}
\bibliography{references}
\newpage
{\color{black}\section*{Appendix}
\subsection{DeltaHand Comparison}
We compare between the adapted hand design for this work and the original design from \deltahands~\cite{si2024deltahands} as shown in Fig.~\ref{fig:compare}. In this work, we decoupled the center actuator and changed 1) the fingertip design, 2) the finger links and joints design, and 3) added an in-hand camera. The new fingertip has an omnidirectional design and an additional soft elastomer coating layer to increase the contact area and contact friction. This has greatly improved grasp stability during experiments. The finger links have embedded rigid PLA cores and the finger joints have smaller gaps to improve the kinematics accuracy and force profile. We observe that the hand can exert stronger forces to manipulate daily objects such as pushing to close a syringe plunger while the original design from \deltahands~\cite{si2024deltahands} can only manipulate lighter objects such as clothes or cables. The decoupled actuators give independent 3 DoF control for each finger to increase dexterity. The inclusion of an in-hand camera provides clear in-hand visual observations for closed-loop policy learning. 

\begin{figure}[t]    
    \centering
    \includegraphics[width=0.98\linewidth]{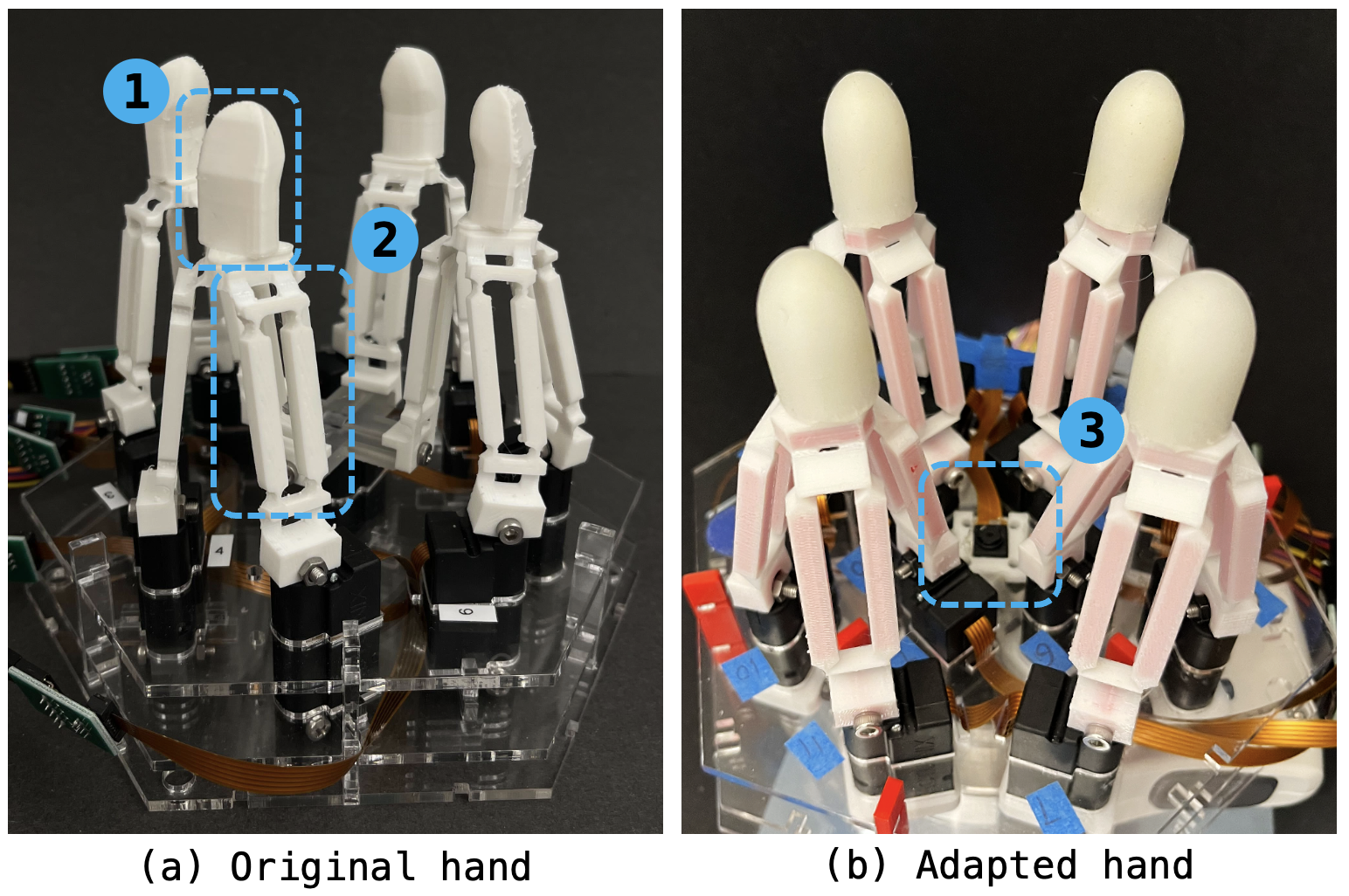}
    \caption{\Wtext{Comparisons between (a) the hand from \deltahands~\cite{si2024deltahands} and (b) our adapted hand. We customize 1) the fingertips to an omnidirectional design for larger contact area with a soft coating elastomer layer for increased friction, 2) the finger links to a rigid core-embedded multi-material design and the finger joints to a better defined shape for reducing kinematics error and increasing force profile, 3) an in-hand camera to provide clear visual observations for closed-loop policy learning.}}
    \label{fig:compare}
    \vspace{-1em}
\end{figure}
\subsection{Teleoperation Interface Details}
\paragraph{Communication and Control from TeleHand to DeltaHand }
The overview of the teleoperation system can be seen in Fig.~\ref{fig:control}. The whole system can be modularized into the Control PC, DeltaHand, TeleHand and external sensors. The Control PC sends and receives all control and sensor signals from the other modules as well as serves as the main computing source for policy inference. The DeltaHand module includes the hand and the in-hand RGB Camera. A microcontroller (Adafruit Feather M0) is used to send and receive the current and desired joint states of the actuators and run a PID control loop. The in-hand camera is connected to a Raspberry Pi which sends images to the Control PC. The TeleHand uses a microcontroller (Arduino Mega) to read and send the sliders' positions when it is teleoperated by a human. Other sensors such as the external RGB camera can be connected directly to the Control PC. All sensor and control signals are published as ROS topics and their frequencies are listed in Table.~\ref{table:control-specs}. As a data pre-processing step for policy learning, we synchronize all the streaming data to the joint states' frequency (20 FPS) by using their ROS timestamps. 

\begin{figure}[t]    
    \centering
    \includegraphics[width=0.98\linewidth]{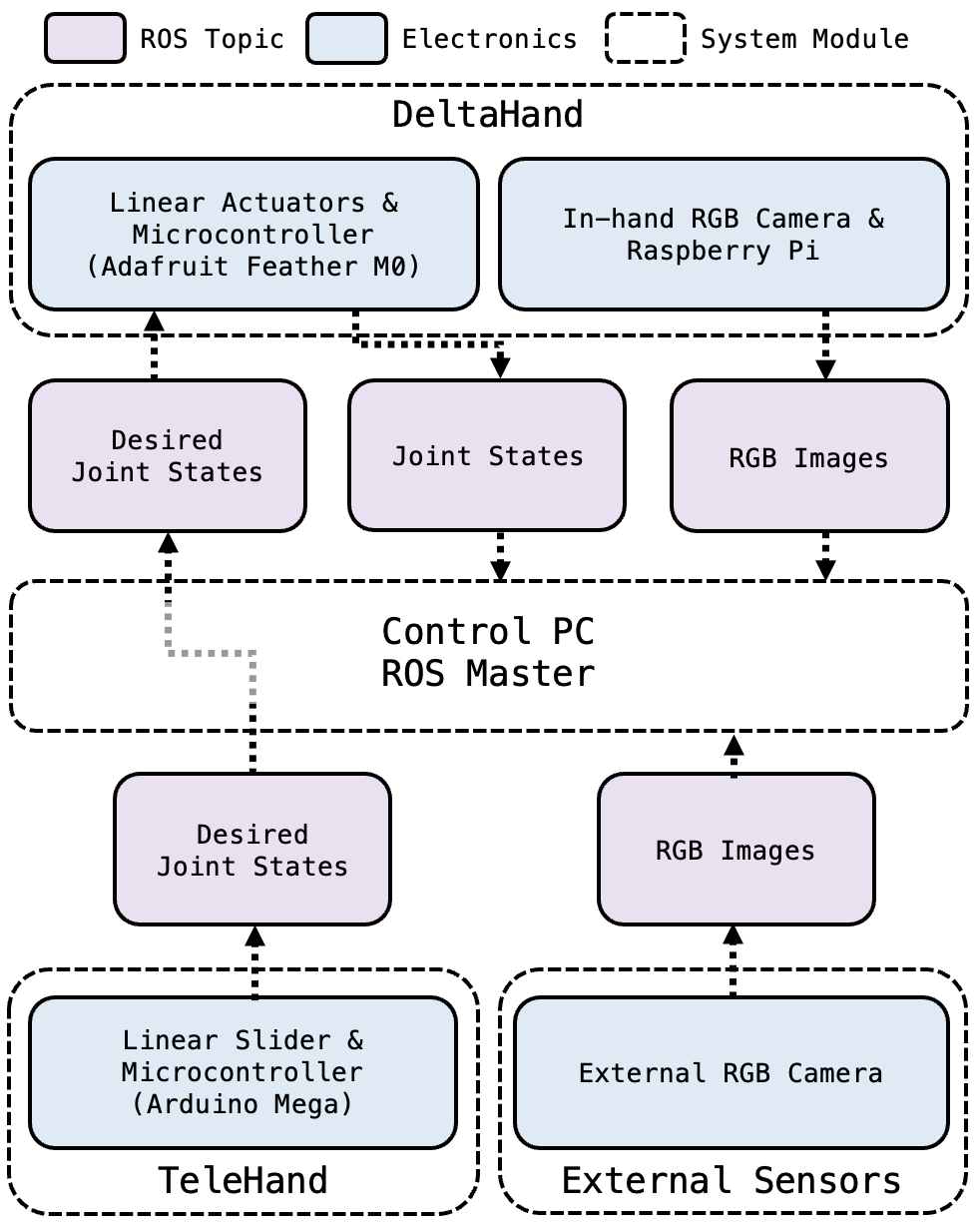}
    \caption{\Wtext{Overview of the teleoperation system. The system can be modularized to the control PC, DeltaHand, TeleHand, and external sensors. The Control PC receives and sends all sensor and control signals as ROS topics. It also serves as the main computing source for policy execution. The DeltaHand has its own microcontrollers for the actuators and sensors. The TeleHand is used to interface with a human to get control signals from the slider potentiometers. All other sensors belong to the external sensors module.}}
    \label{fig:control}
    \vspace{-1em}
\end{figure}

\begin{figure}[t]    
    \centering
    \includegraphics[width=0.98\linewidth]{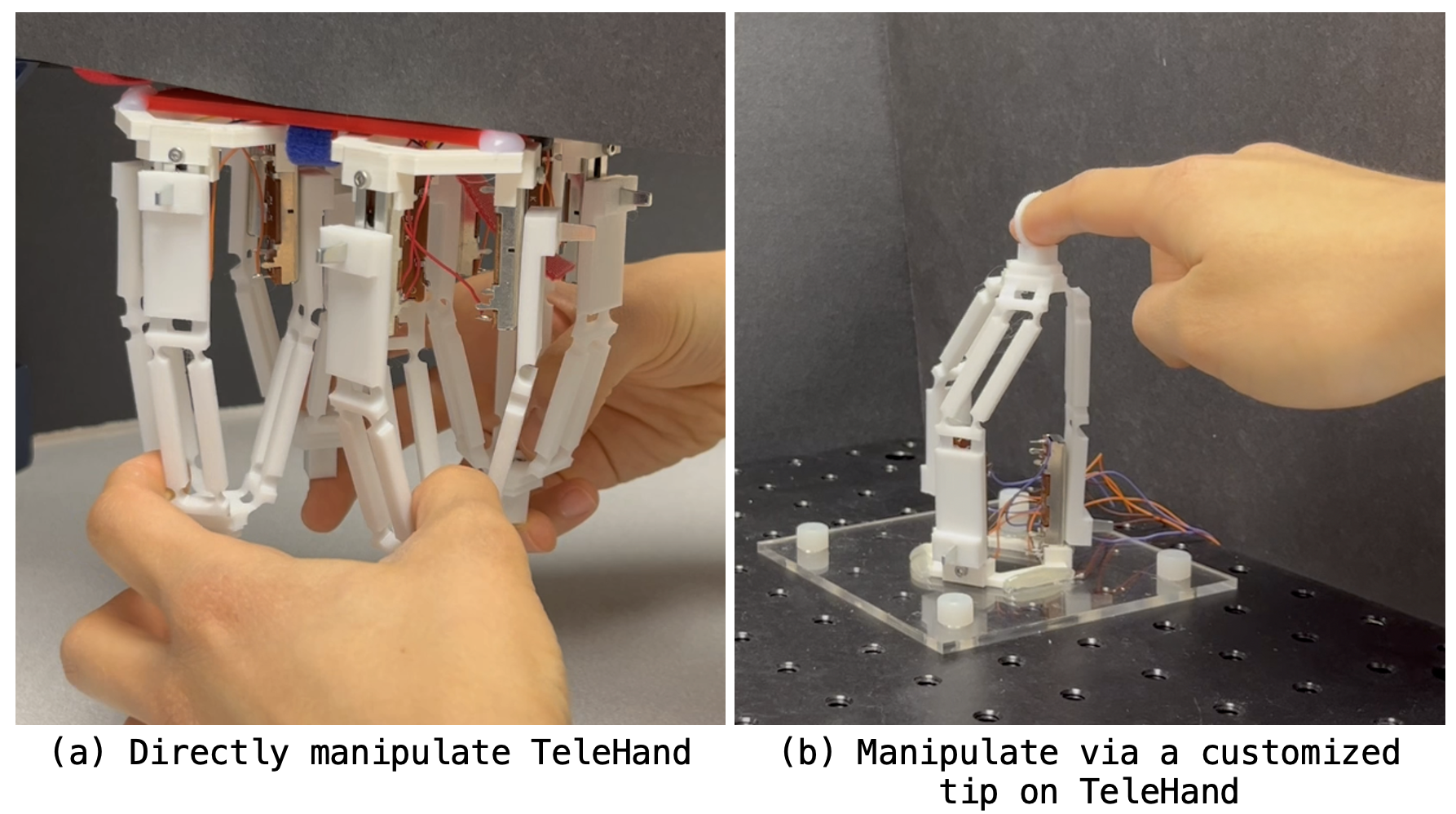}
    \caption{\Wtext{The TeleHand can be teleoperated differently depending on the human's preference. The human can a) directly manipulate the TeleHand by grabbing the links or tips or b) attach and use a customized end-effector such as a ring to constrain their finger to the TeleHand's end-effectors.}}
    \label{fig:telehand-customize}
    \vspace{-1em}
\end{figure}

\begin{table*}[t]
\centering
\setlength{\tabcolsep}{4pt}
\renewcommand{\arraystretch}{1.2}
\begin{tabular}{lccccc}
\Xhline{2\arrayrulewidth}
\textbf{} & \textbf{Joint State} & \textbf{Control (Desired Joint Sate)} & \textbf{Linear Slider} & \textbf{In-hand Camera} & \textbf{External Camera}\\
\hline
{\textbf{Frequency (Hz)}} & 20 &  33  & 133 & 30 & 10 \\
\Xhline{2\arrayrulewidth}
\end{tabular}
\caption{\Wtext{Frequency of sensor and control signals on ROS.}}
\label{table:control-specs}
\vspace{-0.5em}
\end{table*}

\begin{table*}[t]
  \centering
  \small
  \renewcommand{\arraystretch}{1.2}
  \begin{tabular}{p{1.8cm}ccccccc}
    \hline
    {\textbf{Teleoperation}} & {\textbf{Communication}} & \multicolumn{3}{c}{\textbf{Demonstration Collecting Time (seconds)}$\downarrow$} & \multicolumn{3}{c}{\textbf{Mapping Error (mm)}$\downarrow$}\\
    \cmidrule(lr){3-5}\cmidrule(lr){6-8}
    \textbf{Interfaces} & \textbf{Frequency (Hz)}$\uparrow$ & \textbf{Block Slide} & \textbf{Cap Twist} & \textbf{Shape Insert} & \textbf{Block Slide} & \textbf{Cap Twist} & \textbf{Shape Insert} \\
    \hline
    Visual Tracking & 10 & 54.05 & 81.51 & 14.47 & 0.60 & 0.34 & 0.31 \\
    \textbf{Ours} & 133 & 16.99 & 31.11 & 11.81 & 0.23 & 0.23 & 0.20 \\ 
    \hline
  \end{tabular}
  \caption{\Wtext{Teleoperation interface characterization and comparison with visual tracking by using a Leap Motion camera. Our proposed TeleHand has better performance on all metrics when evaluating communication efficiency, demonstration collection time, and mapping errors.}}
  \label{tab:compare-teleop}
  \vspace{-1.4em}
\end{table*}

\begin{figure*}[t]    
    \centering
    \includegraphics[width=0.95\linewidth]{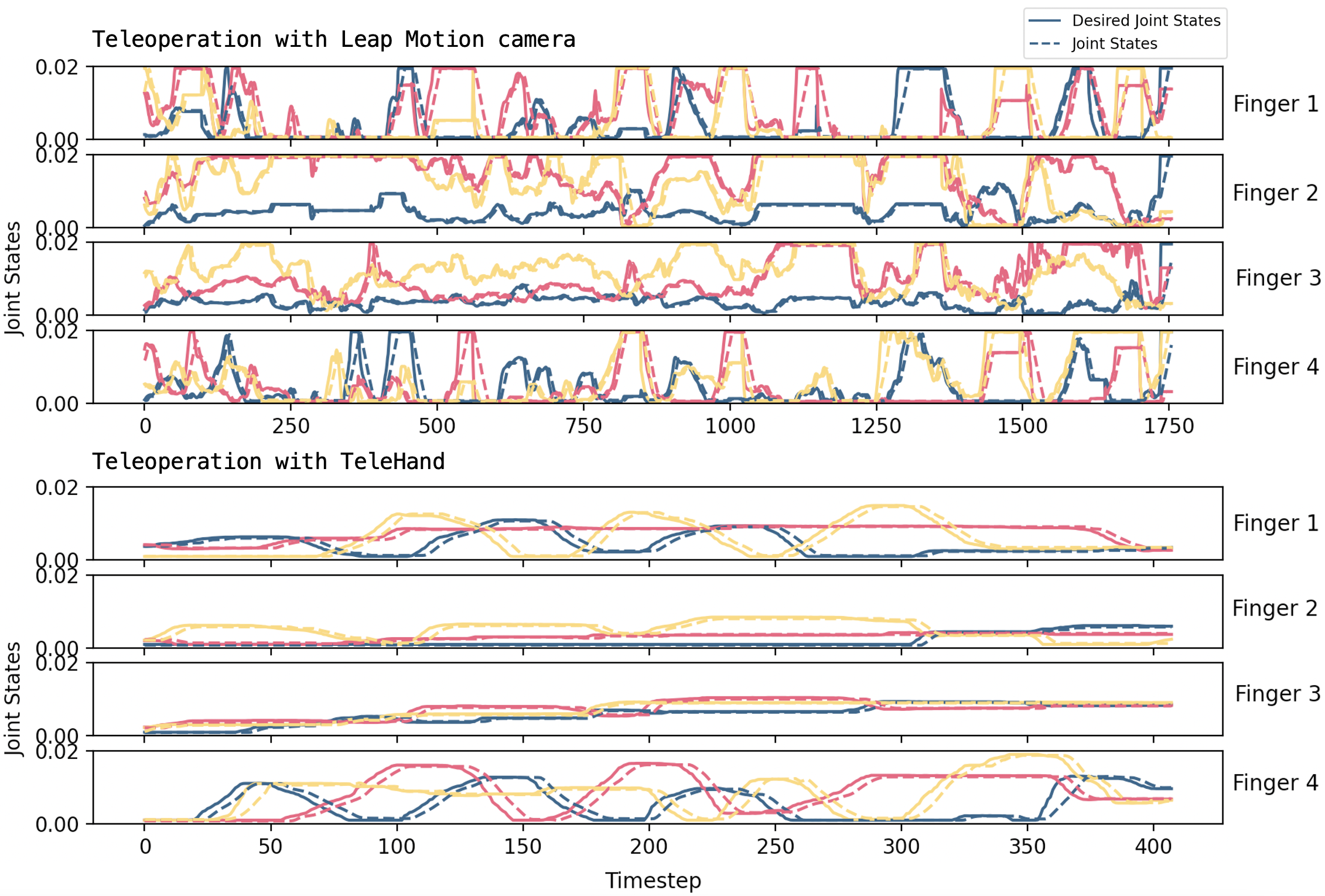}
    \caption{\Wtext{Visualization of the desired joint states and joint states of the DeltaHand when being teleoperated with a Leap Motion camera or TeleHand. We observe less noise and simpler motions from teleoperation with the TeleHand. With direct one-to-one joint mapping, the TeleHand does not suffer from kinematics constraints, which are shown as saturated joint states when teleoperating using the Leap Motion.}}
    \label{fig:teleop-compare}
    \vspace{-1.1em}
\end{figure*}

\paragraph{TeleHand Customization}
As the DeltaHand's kinematic twin, TeleHand is also modularized and its fingertips can be customized in different ways as shown in Fig.~\ref{fig:telehand-customize}. The human user can directly manipulate the TeleHand by grabbing its links or tips. Alternatively, the human's fingers can be more firmly attached to the TeleHand's end-effectors by adding customized tips such as rings or finger gloves. 
With direct manipulation, it is easy to attach and detach the human fingers from the TeleHand while keeping the end-effectors of the TeleHand in the same position due to friction from the sliders. In this manner, human teleoperators can easily reposition their hands on the TeleHand for different motions, or even regrasp the TeleHand to only manipulate a subset of fingers while the others remain stationary. With additional customized fingertips, there's a clear human finger to TeleHand finger to DeltaHand finger motion mapping. Through our experiments, we found that directly manipulating TeleHand is the easiest and most stable way for collecting demonstrations because humans are good at using hand synergies such as squeezing or spreading the four fingers along a desired axis using two hands. 

\paragraph{Teleoperation Interface Comparison}
Visual tracking~\cite{handa2020dexpilot, arunachalam2023dexterous, shaw2023leap} has broadly been used for teleoperating anthropomorphic robotic hands. We characterize and compare two teleoperation interfaces: the TeleHand and visual tracking by using a Leap Motion camera as presented in \deltahands~\cite{si2024deltahands} on three metrics: communication efficiency, human demonstration collection time, and mapping errors from the teleoperation inputs to the robotic hand motions. We evaluate these metrics using three tasks: \textbf{Block Slide}, \textbf{Cap Twist}, and \textbf{Shape Insert}. For each task, we collect three demonstrations until task completion. We calculate the average performance over these three runs and report the results in the Table~\ref{tab:compare-teleop}. 

From the results, we observe that our proposed teleoperation interface, TeleHand, has better performance on all metrics. In addition, we visualize desired joint states and joint states from a demonstration of \textbf{Block Slide} in Fig.~\ref{fig:teleop-compare}. We observe that by using the TeleHand, the signals are less noisy and the motions are simpler which helps policy learning. We also observe saturated joint states during teleoperation using the Leap Motion camera. This is because the human teleoperator may move their fingers outside the workspace of the DeltaHand which leads to an invalid inverse kinematics solution and a noisy trajectory as the human needs time to move their finger back within the workspace. Our TeleHand has one-to-one joint mapping and does not experience this issue.

\subsection{Comparison with Existing Work}
We compare our proposed system with state-of-the-art work from two aspects: a teleoperation system based on hardware development (Table.~\ref{tab:teleop-compare}) and a learning system based on policy learning for dexterous robotic hands (Table.~\ref{tab:learning-compare}). For teleoperation system comparison, we show that our system has a relatively low cost while preserving high dexterity. The compliance of the soft links and fingertips of the DeltaHand also provides higher tolerance for policy deviations as compared to rigid hands. We plan to open-source our system to provide accessibility to potential users. For policy learning comparison, we show that our system has demonstrated capabilities on various tasks including grasping, translational, and rotational tasks. By leveraging end-to-end real-world imitation learning, policy learning is more data efficient and has less data distribution shifts. Our system also provides proprioceptive and visual feedback for closed-loop control.

\begin{table*}[t]
  \centering
  \small
  \renewcommand{\arraystretch}{1.2}
  \begin{tabular}{p{2.0cm}ccccccc}
    \hline
    {{\textbf{Teleoperation}}} & {\textbf{Robotic}} & \textbf{Cost} & \textbf{\# DoF} & \textbf{Hand} & \textbf{Hand} & \textbf{Teleoperation} & \textbf{Availability} \\
    \textbf{System} & \textbf{Hand} & & & \textbf{Type} & \textbf{Material} & \textbf{Interface} & \textbf{}\\
    \hline
    \multirow{2}{1.5cm}{DexPilot~\cite{handa2020dexpilot}} & Allegro & \$15,000 & 16 & Anthropomorphic & Rigid & Vision & Off-the-shelf\\
    & & & & & & (RealSense) & \\
    \multirow{2}{1.5cm}{DIME~\cite{arunachalam2023dexterous}} & Allegro & \$15,000 & 16 & Anthropomorphic & Rigid & Vision & Off-the-shelf\\
    & & & & & & (RealSense) & \\
    \multirow{2}{1.5cm}{LEAP Hand~\cite{shaw2023leap}} & LEAP Hand & \$2,000 & 16 & Anthropomorphic & Rigid & Vision (RGB Camera) & Open-sourced \\
    & & & & & & Manus Meta Glove & \\
    {Stewart Hand~\cite{mccann2021stewart}} & Stewart Hand & \$600 & 6 & Underactuated & Rigid & Space Mouse & Open-sourced \\
    {~\cite{mannam2023framework}} & DASH Hand & \$1500 & 16 & Anthropomorphic & Soft & Manus Meta Glove & Open-sourced \\
    {~\cite{rbovisionteleop}} & RBO Hand 3 & \$2350* & 16 & Anthropomorphic & Soft & Vision (Webcam) & Open-sourced \\
    {\textbf{Ours}} & DeltaHand & \$1,000 & 12 & Exactly Constrained & Soft & Kinematic Twin & Open-sourced\\ 
    \hline
  \end{tabular}
  \caption{\Wtext{Comparison with other state-of-the-art teleoperation systems for dexterous robotic hands. We show that our system has a relatively lower cost while still preserving high dexterity. The DeltaHand has soft fingertips and a compliant finger structure which assists with adapting to different objects and environments and tolerating deviations from learned policies. *The estimated cost of the RBO Hand 3 is \$250 for manufacturing the hand, \$480 for 16 Freescale MPX4250 pressure sensors, and \$1600 for 16 pneumatic Matrix Series 320 valve controllers based on \cite{deimel2016mass, puhlmann2022rbo}.}}
  \label{tab:teleop-compare}
\end{table*}

\begin{table*}[t]
  \centering
  \small
  \renewcommand{\arraystretch}{1.2}
  \begin{tabular}{p{1.5cm}cccccccc}
    \hline
    {\textbf{Learning}}  & \textbf{Robotic} & \multicolumn{3}{c}{\textbf{Task Types}} & \multicolumn{2}{c}{\textbf{Policy Learning}} & \multicolumn{2}{c}{\textbf{Feedback}}\\
    \cmidrule(lr){3-5}\cmidrule(lr){6-7}\cmidrule(lr){8-9}
    \textbf{System} & \textbf{Hand} & \textbf{Grasping} & \textbf{Translational} & \textbf{Rotational} & \textbf{Method} & \textbf{Environment} & \textbf{Proprioception} & \textbf{Vision} \\
    \hline
    DIME~\cite{arunachalam2023dexterous} & Allegro &  &  & \checkmark & IL & Sim \& Real & \checkmark  & \checkmark \\
    \multirow{2}{1.5cm}{LEAP Hand~\cite{shaw2023leap}} & LEAP Hand & \checkmark &  & \checkmark & RL & Sim-to-Real & \checkmark  & \\
    & & & & & & & &\\
    \multirow{2}{1.5cm}{Visual Dexterity~\cite{chen2023visual_dexterity}} & D'Claw~\cite{ahn2020robel} & &  & \checkmark & RL & Sim-to-Real & \checkmark & \checkmark\\
    & & & & & & & &\\
    DEFT~\cite{kannan2023deft} & DASH Hand~\cite{mannam2023framework} & \checkmark &  & \checkmark & RL & Real & & \checkmark \\
    \multirow{2}{1.5cm}{~\cite{patidar2023hand}} & RBO Hand 3~\cite{puhlmann2022rbo} &  & \checkmark & \checkmark & Motion & Real & & \\
    & & & & & Planning & & &\\
   ~\cite{akkaya2019solving} & Shadow Hand~\cite{shadowhand} &  &  & \checkmark & RL & Sim-to-Real & & \checkmark \\
    \multirow{2}{1.5cm}{~\cite{morgan2022complex}} & Yale Model Q~\cite{7090666} & \checkmark & \checkmark & \checkmark & Motion & Real &  & \checkmark\\
    & & & & & Planning & & &\\
    \textbf{Ours} & DeltaHand~\cite{si2024deltahands} & \checkmark  & \checkmark  & \checkmark  & IL & Real & \checkmark  & \checkmark \\ 
    \hline
  \end{tabular}
  \caption{\Wtext{Comparison with other state-of-the-art policy learning systems for dexterous robotic hands. We show that our system is evaluated on a variety of tasks including grasping, in-hand translation and rotation tasks. Leveraging end-to-end real-world imitation learning has the benefits of data efficiency and less data distribution shift.}}
  \label{tab:learning-compare}
\end{table*}

\subsection{Learning Policy Implementation Details}
\label{sec:policy-detail}
For an ablation study on using different imitation learning methods, we adapted the implementation of Behavior Cloning (BC), Implicit Behavior Cloning (IBC), and Diffusion Policy from~\cite{chi2023diffusion, zakka2021ibc}. For fair comparison, we use the same ResNet18 image encoder adapted from Diffusion Policy~\cite{chi2023diffusion} for all three methods. For BC and IBC, we use the same 5-layer MLP network architecture consisting of \{$N_{input}$,1024,1024,512,256,$N_{output}$\} perceptrons, where $N_{input}$ and $N_{output}$ are the number of input and output channels. For BC,  $N_{input} = (512 + 12) * 2$, where 512 is the dimension of the image feature, 12 is dimension of the joint state, and 2 is the observation horizon; $N_{output} = 12 * 8$, where 12 is the action dimension, and 8 is the execution horizon. For IBC, $N_{input} = (512 + 12) * 2 + 12 * 8$, where 512 and 12 are again the dimensions of the image feature and joint state respectively, 2 is the observation horizon, and 8 is the execution horizon; $N_{output} =1 $, is the energy of the observation-action pair. We use the same dataset, data normalization, data augmentation, training configurations including optimizer configuration, and number of training epochs for BC and IBC that we used to train the diffusion policy. For IBC training, we sample 256 random actions for each observation input; and during inference, we iteratively sample 1024 random actions 3 times with a noise scale starting from 0.33 and then reducing it by a factor of 0.5 in each iteration.

}

\end{document}